\documentclass[10pt,twocolumn,letterpaper]{article}

\usepackage[pagenumbers]{style} 

\usepackage[T1]{fontenc}
\usepackage{url}
\usepackage{xspace}
\usepackage{graphicx}
\usepackage{booktabs}       
\usepackage{wrapfig}
\usepackage{enumitem}

\definecolor{analysisblue}{RGB}{31,119,180}
\definecolor{analysisorange}{RGB}{255,127,14}
\definecolor{analysispurple}{RGB}{123,104,238}

\newcommand{\ourmethod}{\textsc{MADrive}\xspace}
\newcommand{\ourdataset}{\textsc{MAD-Cars}\xspace}

\newcommand{\app}[1]{Appendix~\ref{appendix:#1}}
\newcommand{\fig}[1]{Figure~\ref{fig:#1}}
\newcommand{\tab}[1]{Table~\ref{tab:#1}}
\newcommand{\sect}[1]{Section~\ref{sect:#1}}

\definecolor{blue}{rgb}{0.21,0.49,0.74}
\usepackage[pagebackref,breaklinks,colorlinks,allcolors=blue]{hyperref}

\title{MADrive: Memory-Augmented Driving Scene Modeling}

\author{Polina Karpikova\thanks{Equal contribution.}\;\;\thanks{Email: p.karpikova@yandex.ru} \\
Yandex 
\and
Daniil Selikhanovych$^{*}$ \\
Yandex Research, HSE University 
\and
Kirill Struminsky$^{*}$ \\
Yandex Research, HSE University 
\and
Ruslan Musaev \\
Yandex 
\and
Maria Golitsyna \\
Yandex 
\and
Dmitry Baranchuk \\
Yandex Research 
}

\begin{document}
\maketitle

\begin{abstract}
Recent advances in scene reconstruction have pushed toward highly realistic modeling of autonomous driving (AD) environments using 3D Gaussian splatting. 
    However, the resulting reconstructions remain closely tied to the original observations and struggle to support photorealistic synthesis of significantly altered or novel driving scenarios. This work introduces \ourmethod, a memory-augmented reconstruction framework designed to extend the capabilities of existing scene reconstruction methods by replacing observed vehicles with visually similar 3D assets retrieved from a large-scale external memory bank. 
    Specifically, we release \ourdataset, a curated dataset of ${\sim}70$K 360° car videos captured in the wild and present a retrieval module that finds the most similar car instances in the memory bank, reconstructs the corresponding 3D assets from video, and integrates them into the target scene through orientation alignment and relighting.
    The resulting replacements provide complete multi-view representations of vehicles in the scene, enabling photorealistic synthesis of substantially altered configurations, as demonstrated in our experiments.
    
    \vspace{1mm}
    \centering {\small \textbf{Project page:} 
    
    \url{https://yandex-research.github.io/madrive/}}
\end{abstract}

\vspace{-6mm}
\begin{figure*}[!ht]
  \hspace{-2mm}
  \includegraphics[width=1.0\textwidth]{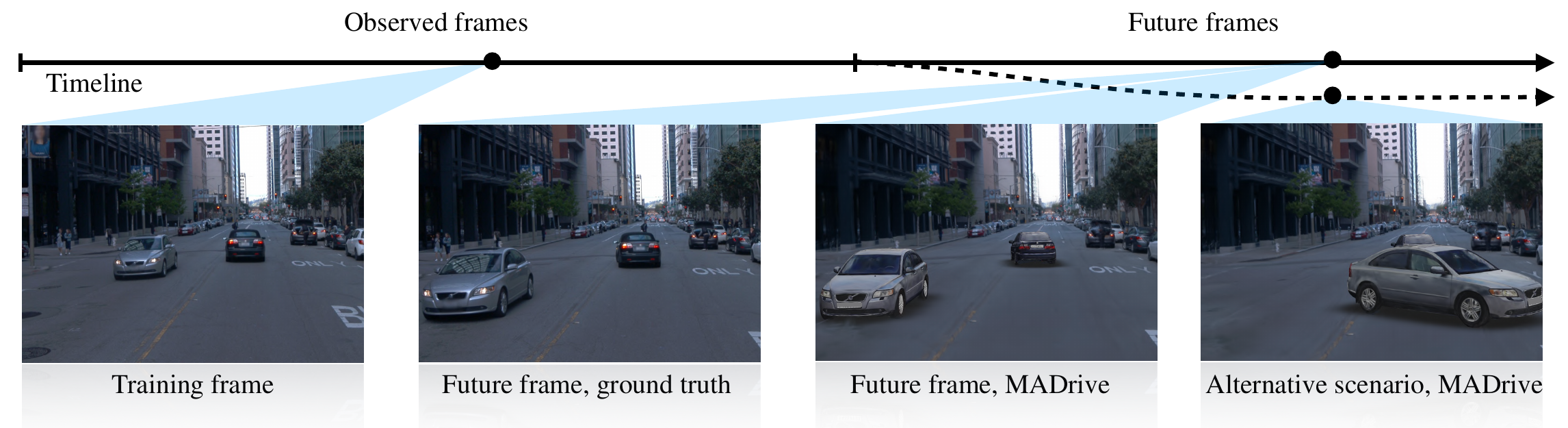}
  \vspace{-2mm}
  \caption{
      \textbf{\ourmethod} reconstructs a 3D driving scene from training frames ({\color{analysisblue}Left}) and replaces partially observed vehicles in the scene with realistically reconstructed counterparts retrieved from \textbf{\ourdataset}, our novel multi-view auto dataset.
      \ourmethod enables high-fidelity modeling of future scene views ({\color{analysisorange}Middle-left vs. Middle-right}) and supports simulation of alternative scenarios, advancing novel-view synthesis in dynamic environments ({\color{analysispurple}Right}).
  }
  \vspace{-2mm}
  \label{fig:teaser}
\end{figure*}

\section{Introduction}

Modern autonomous driving (AD) systems heavily rely on computer vision and machine learning models trained on large-scale and diverse datasets~\citep{xiao2021pandaset, sun2020scalability, geiger2012areweready, cabon2020virtual, caesar2020nuscenes}.
However, collecting and annotating such data in the real world is expensive, time-consuming, and often limited by safety and practicality.
Driving simulators~\citep{wang2023cadsim, zhou2024hugsim, yang2023unisim} offer an alternative by generating realistic novel views and rare, safety-critical scenarios that are difficult to record in the real world.
Realistic simulation helps reduce the domain gap between synthetic and real data and improves model robustness.
Moreover, controllable reenactments of driving scenes also allow systematic testing of perception and planning models by reproducing failure cases under different conditions.
In this work, we reconstruct driving sequences captured by autonomous vehicle to enable such controllable, photorealistic reenactments for safety testing and data augmentation.

Recent progress in multi-view reconstruction and novel view synthesis~\citep{kerbl20233d, yu2024mip, kheradmand20243d} has enabled photorealistic reconstruction of complex real-world scenes, forming a strong foundation for realistic driving simulators~\citep{zhou2024hugsim}.
Unlike game engine–based environments, such methods preserve the visual properties of real data, reducing domain shift.
Modern driving scene reconstruction methods~\citep{zhou2024hugs, yan2024street, khan2024autosplat} can accurately reproduce observed trajectories and support small viewpoint changes.
However, their quality drops when extrapolating far from the observed data, as unseen regions lack geometric and photometric information.
As a result, they cannot reliably simulate large trajectory changes such as U-turns or parking maneuvers, where missing observations cause visible artifacts and incomplete geometry.

To overcome these limitations, we replace dynamic objects with high-quality external assets.
Traditional computer graphics assets are manually created by artists, making it costly to build a large and diverse library.
Instead, we reconstruct assets directly from real-world data.
We introduce \ourdataset, \textbf{M}ulti-view \textbf{A}uto \textbf{D}ataset — a large-scale collection of $360^\circ$ vehicle video videos. 
The dataset contains about 70,000 car instances covering various makes, models, colors, providing diverse and realistic inputs for asset reconstruction.
This approach greatly expands the diversity of assets compared to existing public car datasets.

To reconstruct realistic and reusable assets, we develop a car reconstruction pipeline tailored for our in-the-wild captures.
A key challenge is adapting models captured under unknown and scene-specific lighting to new environments.
To solve this, we propose a relightable variant of Gaussian splatting that supports physically based rendering under novel illumination.
Our optimization scheme uses a generative model~\citep{wu2025difix3d+} to simulate multi-lighting supervision during reconstruction, effectively separating material properties from lighting.
This enables high-quality relighting and seamless integration of assets into new scenes.

Building on these components, we present \ourmethod, a framework that replicates driving sequences by replacing observed vehicles with visually similar reconstructed 3D assets retrieved from a large external dataset.
Using high-fidelity, relightable assets, \ourmethod preserves realistic appearance even under large trajectory changes.
The framework enables controllable reenactments by rendering future or alternative frames beyond the original sequence.
Quantitative results show that pre-trained perception models perform similarly on our rendered frames and real hold-out data, confirming the realism and consistency of our reconstructions for downstream AD tasks.

In summary, our contributions are threefold: (1) a large-scale dataset of in-the-wild $360^\circ$ vehicle captures for diverse asset reconstruction, (2) a relightable Gaussian splatting pipeline that separates object appearance from illumination for realistic rendering, and (3) an integrated framework that reconstructs, reenacts, and validates driving scenes with controllable trajectories, achieving photorealism and consistency with real-world perception results.

\section{Related Work}
\paragraph{Dynamic Urban Scene Reconstruction.}
Recent dynamic 3D scene reconstruction works adopt 3D Gaussian splatting~\citep{kerbl3Dgaussians} as an efficient and expressive representation~\citep{yang2023real, wu20244d, yang2024deformable, chen2023periodic}.
Several approaches, including StreetGS~\citep{yan2024street}, AutoSplat~\citep{khan2024autosplat}, and HUGS~\citep{zhou2024hugs}, apply this representation to driving scenes by decomposing them into static backgrounds and dynamic vehicles placed with tracked 3D bounding boxes.
HUGS incorporates optical flow and semantic segmentation to guide optimization and adds realistic shadow modeling, while AutoSplat improves vehicle reconstruction by exploiting bilateral symmetry and better initialization from image-to-3D priors~\citep{pavllo2023shape}.
Other works~\citep{zhou2024drivinggaussian, chen2025omnire} introduce dynamic Gaussian graphs to handle multiple moving objects of different nature.
We refer the reader to~\app{related_work_nerf} for a review of earlier NeRF-based scene representation methods.
While these methods improve the fidelity of reconstructed urban scenes, they rely solely on observational data to model dynamic objects, making it difficult to capture complete vehicle geometry and appearance under sparse or occluded views.

\vspace{-2mm}
\paragraph{3D Car Datasets.}
Several public datasets provide 3D car assets. 
Early collections such as SRN-Car~\citep{chang2015shapenet} and Objaverse-Car~\citep{deitke2023objaverse} consist of CAD models that differ notably from real vehicles in texture realism and geometric detail.
More recent efforts~\citep{zhang2021ners, du20243drealcar} have focused on real-captured 3D car datasets.
MVMC~\citep{zhang2021ners} includes $576$ cars, each with an average of $10$ views.
3DRealCar~\citep{du20243drealcar} provides $2,500$ car instances, each with ${\sim}200$ dense high-resolution RGB-D views.
In contrast, \ourdataset includes ${\sim}70,000$ 360° car videos at a comparable resolution and average number of views as 3DRealCar, offering substantially greater generalization and diversity.

\vspace{-2mm}
\paragraph{NVS with External 3D Car Assets.}
HUGSim~\citep{zhou2024hugsim} builds a closed-loop AD simulator by inserting 3D car models from 3DRealCar~\citep{du20243drealcar} into reconstructed scenes.
We instead replace only the vehicles observed in the recording with retrieved matches, producing a close replica of the captured scene for reenactment.

Several approaches leverage CAD models for scene representation~\citep{engelmann2017samp, wang2023cadsim, uy2020deformation, avetisyan2019scan2cad, gumeli2022roca, wei2024editable}, though such models often lack photorealistic textures and accurate geometry.
To improve realism, some methods perform geometry refinement~\citep{uy2020deformation, wang2023cadsim, engelmann2017samp}, while UrbanCAD~\citep{lu2024urbancad} retrieves visually similar CAD models and refines their textures and illumination to better match the scene.
However, the obtained models still have a noticeable gap in realism and correspondence to actual cars. 
In contrast, \ourmethod retrieves real car instances from a large-scale dataset spanning diverse brands, materials, and lighting conditions, helping to narrow this realism gap.





\begin{figure*}
    \centering
    \vspace{-11mm}
    \includegraphics[width=\textwidth]{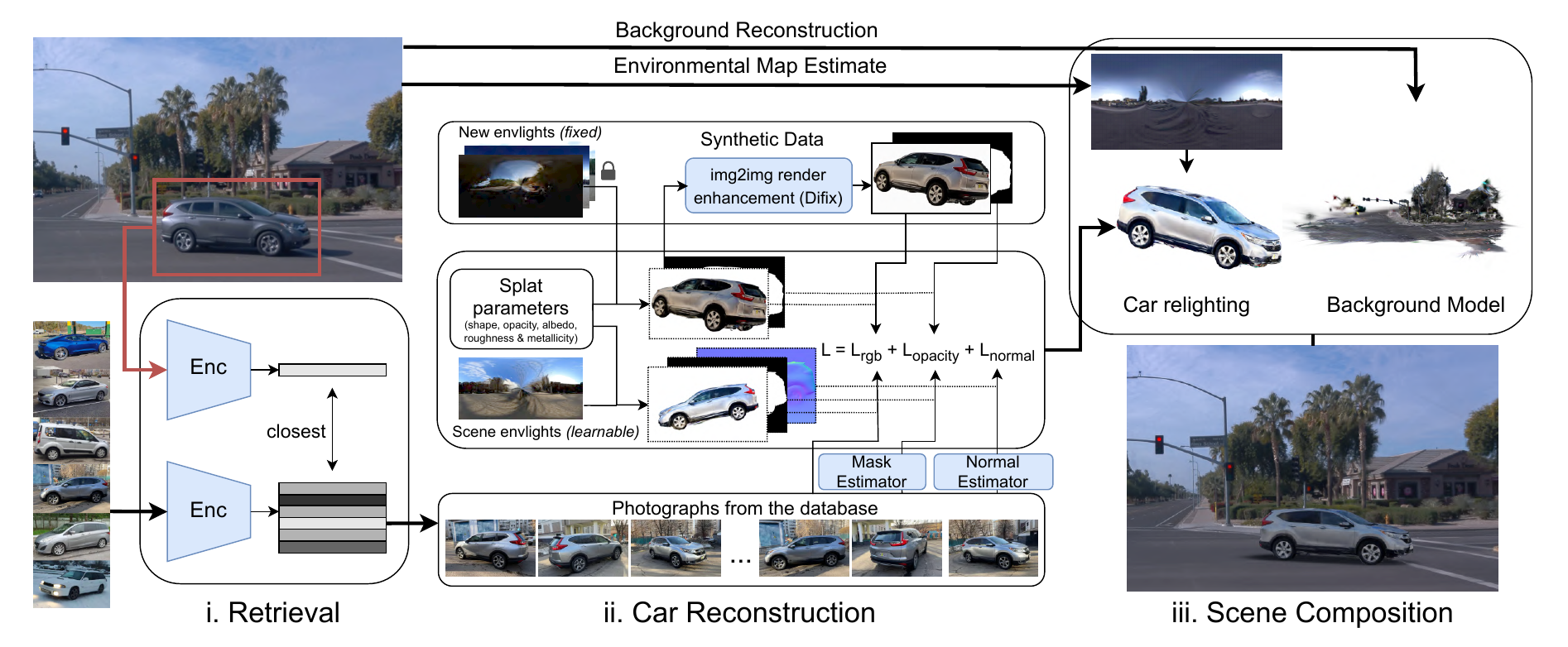}
    \vspace{-8mm}
    \caption{
    \textbf{\ourmethod Overview.}
        Given an input frame sequence, our retrieval scheme finds similar vehicles in an external database ({\color{analysisblue} Left}).
        The 3D reconstruction pipeline then produces detailed vehicle models from the retrieved videos.
        The vehicles are represented using relightable 2D Gaussian splats.
        To enable relighting, we generate synthetic novel views under multiple illumination conditions.
        Opacity masks are applied to remove background splats, and the model geometry is regularized using external normal maps. ({\color{analysisorange} Middle}).
        The reconstructed vehicles are adapted to the scene’s lighting and composed with the background to produce the overall scene representation ({\color{analysispurple} Right}).
    }
    \label{fig:main}
    \vspace{0mm}
\end{figure*}

\vspace{-2mm}
\paragraph{Relighting.}
Scene reconstruction methods based on radiance fields jointly recover geometry and outgoing radiance from multi-view inputs.
Since radiance depends on scene illumination, object appearance changes under different lighting conditions.gt
The outgoing radiance follows the rendering equation~\citep{kajiya1986rendering}, and exact relighting would require full light transport modeling via ray tracing.
Several NeRF-based methods extend radiance fields to simulate light transport and relighting~\citep{verbin2024nerf, zhao2024illuminerf, tang2025rogr, jin2023tensoir}, but they demand substantial computational resources and are not suitable for real-time simulation.
Although several recent works explore efficient ray tracing for Gaussian splats~\citep{xie2024envgs, govindarajan2025radiant, moenne20243d, byrski2025raysplats}, relighting remains outside their scope.

Several methods instead approximate light transport using simplified shading models compatible with rasterization.
LumiGauss~\citep{kaleta2024lumigauss} performs splat-based relighting with spherical harmonics~\citep{ramamoorthi2001efficient}, but requires multi-illumination data and is limited to diffuse surfaces.
GaussianShader~\citep{jiang2024gaussianshader} employs a split-sum approximation~\citep{karis2013real} to enhance specular reflections during reconstruction.
R3DG~\citep{gao2024relightable} improves relighting quality by jointly decomposing materials and illumination from single-lighting captures using an enhanced illumination model.
While our approach also employs a physically based shading model, it augments training with synthetic multi-illumination data to improve relighting. 

To estimate the environmental map for object insertion, DiPIR~\citep{liang2024photorealistic} employs a generative model and a gradient-based procedure.
In constrast, we estimate the environmental map from training frames using DiffusionLight~\citep{phongthawee2024diffusionlight} without additional costly optimization procedures.
Our approach is modular and can be easily adapted to incorporate alternative environmental map estimation methods such as~\citep{liang2025luxdit} if needed.

\vspace{-2mm}
\section{Method}
In this section, we describe \ourmethod that replaces the vehicles in the scene with visually similar, fully-observed 3D car assets, thereby enabling the prediction of future vehicle appearances following sharp turns or other complex maneuvers.
The overview of our method is presented in Figure~\ref{fig:main}.

\subsection{Scene Decomposition and Reconstruction Overview}
Following~\citep{yan2024street}, we decompose each driving scene into static and dynamic components.
The static part represents the background—ground, surroundings, and sky—while the dynamic part contains all moving vehicles.
The static component can be reliably reconstructed from the ego-vehicle’s motion and depth sensor data, which provide sufficient parallax to recover 3D structure.

We adapt the approach of Street Gaussians~\citep{khan2024autosplat} to model the static scene.
The surroundings are parameterized using 3D Gaussian splats~\citep{kerbl20233d}, the ground is represented by horizontal 2D splats, and the sky is placed at an infinite distance and blended during rendering to avoid depth ambiguities.

Dynamic elements include all moving vehicles; however, we treat cars marked as stationary in the metadata as part of the static background for simplicity.
Modeling dynamic objects poses two main challenges: handling compound motion and reconstructing objects from incomplete observations (e.g., when only one side of a vehicle is visible).
Following prior work~\citep{yan2024street, khan2024autosplat}, we approximate each observed vehicle as a set of static Gaussian splats within its moving 3D bounding box, effectively capturing its motion during training.
Both static and dynamic components are initialized from LiDAR points and jointly optimized using a photometric loss.

During inference, we reuse the static part of the scene.
At the same time, we replace moving vehicles with 3D car models extracted from a bank of cars using the retrieval-based approach, described in the next section.
This substitution allows obtaining high-quality renders for configurations diverging from the ones observed during training. The computational bottlenecks of our method are static scene reconstruction ($\approx$3 hours on a single GPU) and per-asset reconstruction ($\approx$30 minutes). Retrieval, asset insertion, relighting, and rendering add only minor overhead and run efficiently in comparison.

\subsection{Database Retrieval}
Our goal is to replace observed vehicles in driving sequences with visually similar 3D assets retrieved from an external database.
This subsection outlines our retrieval pipeline and the corresponding database construction.

\paragraph{Retrieval Query Information.}
For each driving sequence, we project the 3D bounding box of every detected vehicle onto the image plane to obtain a segmentation mask.
After discarding small and overlapping masks, we extract cropped images centered on individual cars.
Each crop is embedded using SigLIP2~\citep{tschannen2025siglip}, while the vehicle color is estimated with Qwen2.5-VL~\citep{yang2024qwen2}.
The color cue complements the embedding features, which primarily capture brand and type information, improving retrieval accuracy.

\begin{figure*}[t]
    \centering
    \vspace{-6mm}
    \includegraphics[width=\textwidth]
    {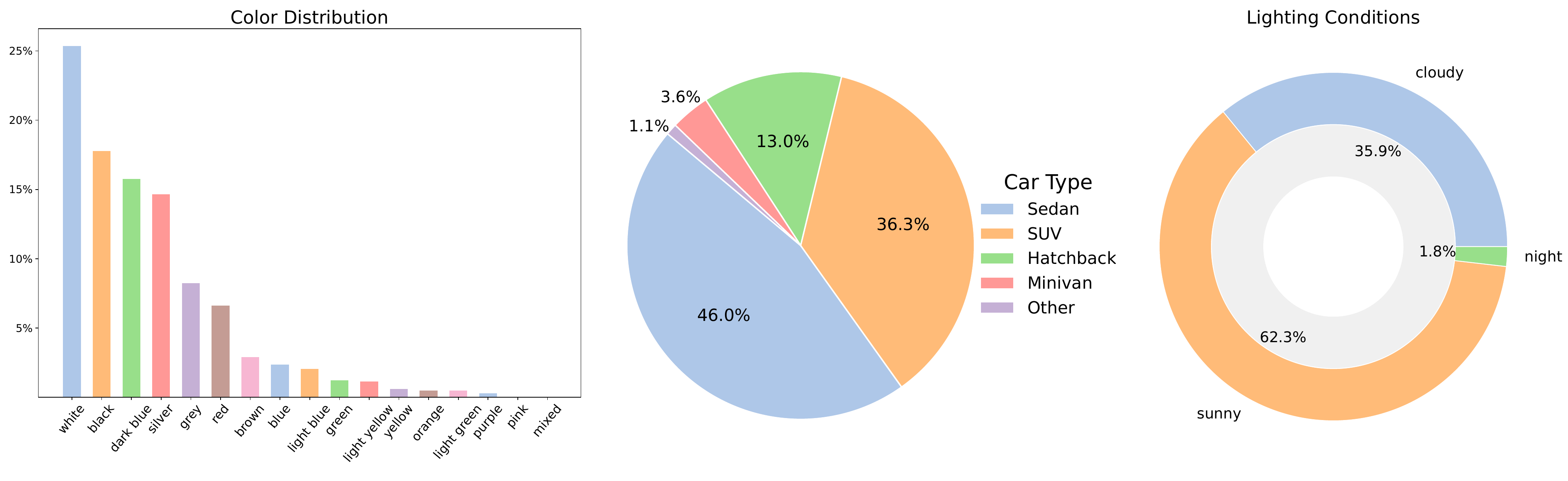}
    \vspace{-6mm}
    \caption{\textbf{\ourdataset Analysis.} Memory-bank statistics on colors ({\color{analysisblue} Left}), car types ({\color{analysisorange} Middle}) and lighting conditions ({\color{analysispurple} Right}).}
    \label{fig:dataset}
    \vspace{-4mm}
\end{figure*}

\paragraph{Database Collection and Statistics}
We introduce \ourdataset, a large-scale collection of in-the-wild multi-view car videos sourced from online car-sale advertisements.
The dataset contains approximately ${\sim}70,000$ car instances, each with about ${\sim}85$ frames at $1920{\times}1080$ resolution.
It spans roughly $150$ brands and covers diverse colors, car types, and various lighting conditions.
Figure~\ref{fig:dataset} summarizes the distributions of color, type, and illumination.
Each instance includes metadata describing car attributes.

To ensure high-quality reconstruction, we curate the dataset by filtering out frames and instances that degrade multi-view consistency.
Specifically, we remove low-quality or overly dark frames using CLIP-IQA~\citep{wang2022exploring}, and employ Qwen2.5-VL~\citep{yang2024qwen2} to detect persistent occlusions that obscure parts of the vehicle across most views—such as nearby grass, bushes, or fences—as well as to filter out interior and obstructed shots.
Further data-collection details are provided in \app{data_collection}.

To retrieve a matching asset, we first pre-select candidates with similar color and then identify the closest match in the embedding space.
The selected instance is reconstructed into a 3D model using its associated multi-view image set.
The reconstruction pipeline is detailed in the following section.

\subsection{Car Reconstruction Deatails}
\paragraph{Relightable Car Models.}
We begin by specifying the representation used to model vehicles.
By default, Gaussian splatting models the entangled radiance observed in the training frames, implicitly coupling surface reflectance and illumination.
In our setup, however, we need to explicitly separate lighting and material effects to enable model insertion into environments with different illumination.
To this end, we adopt a relighting strategy based on physically-based shading~\citep{burley2012physically}.

We use a two-dimensional modification of Gaussian splats~\citep{huang20242d}, which approximates the 3D model with a collection of flat Gaussian splats.
Each splat is parameterized by its location $\mu \in \mathbb R^3$, orientation matrix $R \in SO(3)$, transparency $\alpha \in \mathbb R$, and \textit{two} scale parameters $\sigma_x, \sigma_y \in \mathbb R$.
Unlike 3D splats, 2D splats have well-defined surface normals ${\mathbf n} = {\mathbf n}(R)$, which are essential for surface relighting.

To disentangle scene lighting from surface materials, we adopt the lighting model from~\citep{munkberg2022extracting} for each splat.
The model assumes distant illumination with incident radiance $L_i(\omega_i)$ and defines the outgoing radiance in direction $\omega_o$ according to the rendering equation~\citep{kajiya1986rendering}:
\begin{equation}\label{eq:lighting}
L(\omega_o) = \int_\Omega L_i(\omega_i) f(\omega_i, \omega_o) (\omega_i \cdot {\mathbf n}) \mathrm d \omega_i,
\end{equation}
where $f(\omega_i, \omega_o)$ is the surface BSDF and the integration is taken over the hemisphere $\Omega$ around the surface point. 
The environment lighting $L_i$ is parameterized as a high-resolution cubemap.
Following~\citep{munkberg2022extracting}, we parameterize each splat’s BSDF using the Cook–Torrance shading model~\citep{cook1982reflectance}, with appearance defined by albedo $c \in \mathbb R^3$, roughness $r \in \mathbb R$, and metallicity $m \in \mathbb R$.

Finally, to avoid the cost of directly evaluating Eq.~\ref{eq:lighting}, we employ the differentiable split-sum approximation from~\citep{munkberg2022extracting}, which allows us to jointly infer incident radiance and splat BSDF parameters during optimization.

\paragraph{Reconstruction Algorithm.} Next, we specify the details of the reconstruction algorithm used for the representation above.

For a rendered frame $I_i$ and the ground truth frame $\hat{I}_i$, our objective consists of image-based loss 
$\mathcal{L}_{\text{rgb}} = \mathcal{L}_1(I_i, \hat{I_i}) + \mathcal{L}_{SSIM}(I_i, \hat{I}_i)$ along with several regularizers.
To exclude unnecessary background objects from the model, we generate masks $\hat{M}_i(x, y) = 
[ \hat{I}_i(x, y) \textit{ is part of a car} ]$ with Mask2Former \citep{cheng2022maskedattention} to indicate pixels that belong to the model.
Our opacity loss promotes high transparency outside of car pixels $\mathcal{L}_{\textit{opacity}} = \sum_{x, y}  (1 - \hat{M}_i) \cdot T_i$, where $T_i$ is the transparency map of the rendered frame.
In our model, proper relighting requires accurate surface normals, so we additionally estimate normal maps $\hat{N}_i = n(\hat{I}_i)$ with a NormalCrafter model \citep{bin2025normalcrafter} and use the estimates to regularize Gaussian orientations.
For the rendered normal maps $N_i$, the regularizer is $\mathcal{L}_{\textit{normal}} = \sum_{x, y}  \hat{M}_i \cdot (1 - N_i^T \hat{N_i})$.
The resulting objective is
\vspace{-1mm}
\begin{align}\label{eq:objective}
\begin{split}
  \mathcal{L}_{\textit{gt}}(I_i, \hat{I}_i) = \mathcal{L}_{\textit{rgb}}(I_i, \hat{I}_i) & + \lambda_{\textit{opacity}} \mathcal{L}_{\textit{opacity}}(I_i, \hat{I}_i) + \\
  & \lambda_{\textit{normal}} \mathcal{L}_{\textit{normal}}(I_i, \hat{I}_i).
\end{split}
\end{align}

To address the limited lighting variability in our real captures, we additionally introduce a generative augmentation strategy that extends the Difix framework~\citep{wu2025difix3d+} beyond its original purpose.
While Difix was originally proposed to enhance the photorealism of rendered novel views, we repurpose it to simulate appearance under diverse illumination conditions, effectively approximating multi-illumination supervision.

Concretely, we render the reconstructed model from random novel viewpoints and relight each render using randomly sampled environmental maps.
These low-quality renders are then refined with Difix to produce synthetic enhanced frames $\tilde{I}_i$ that emulate novel realistic lighting.
By pairing each $\tilde{I}_i$ with a physically based render $I_i$ under the same illumination, we compute the objective in Eq.~\ref{eq:objective}, using the $\alpha$-mask of $\tilde{I}_i$ in $\mathcal{L}_{\textit{opacity}}$ and omitting $\mathcal{L}_{\textit{normal}}$ term.

This augmentation effectively disentangles illumination and material properties, allowing our model to generalize across lighting conditions that were never observed in the training data.
We mix synthetic and real frames in equal proportion, introducing synthetic samples after the initial 10k iterations and refreshing them every 2.5k steps for 20k additional iterations.

After the Gaussian splatting reconstruction, we apply several postprocessing steps to obtain the final car asset.
We remove stray splats that do not belong to the depicted car by computing instance segmentation~\citep{yolo11_ultralytics} and discarding those that lie behind the training cameras or project outside the car mask in most views, using a soft threshold to handle segmentation errors.
The cleaned point cloud is then oriented along its principal components, and the front direction is estimated using an orientation model~\citep{scarvelis2024orient}.

\begin{figure*}[t]
    \vspace{-4mm}
    \centering
        \includegraphics[width=\linewidth, page=7]{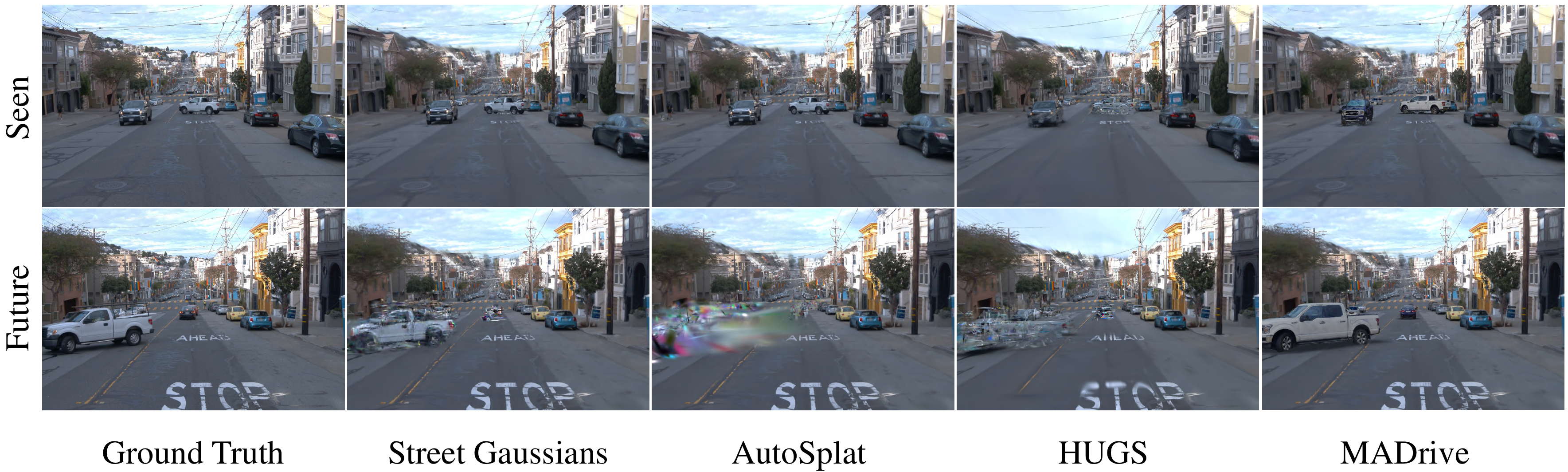}
    \vspace{-6mm}
    \caption{
        \textbf{Qualitative comparison} of \ourmethod with non-retrieval-based driving scene reconstruction methods.
        Reconstruction of the training views ({\color{analysisblue}Top}). 
        Reconstruction of the hold-out (future) views ({\color{analysisorange} Bottom}).
    }
    \vspace{-3mm}
    \label{fig:qualitative_frame_comparison}
\end{figure*}

\subsection{Car Insertion and Relighting}
After reconstructing the vehicle models with a standardized orientation, we integrate them into the captured scene.
We first estimate each car’s orientation by aligning the asset’s bounding box with the detected vehicle bounding box in the scene.
However, directly matching the bounding boxes in scale and position often results in noticeable misalignment.
To obtain accurate placement, we refine the transformation using the Iterative Closest Point (ICP) algorithm~\citep{icp}, aligning the reconstructed asset to the corresponding LiDAR point cloud and deriving the final scale and location from this alignment.

To ensure consistent lighting between the inserted assets and the reconstructed scene, we estimate the scene’s environmental map to compute the outgoing radiance of each asset in Eq.~\ref{eq:lighting}.
Since the Waymo dataset~\citep{sun2020scalability} lacks full $360^\circ$ camera coverage and is captured in low dynamic range, we approximate the full high dynamic range environmental map with DiffusionLight~\citep{phongthawee2024diffusionlight} using the last training frame from the frontal camera.
We further scale the estimated environmental map to minimize tone discrepancies between the last training frame and the rendered image.
Finally, to enhance visual realism, we add a shadow beneath the car, modeled as a black semi-transparent plane composed of 2D splats positioned right below the wheels.


\section{Experiments}
In the following, we present the evaluation of \ourmethod.
~\sect{exp_setup} describes the experimental setup, followed by the main results in~\sect{exp_main_results}.
Finally, ~\sect{ablation_study} analyzes the retrieval, car reconstruction components, and relighting in the scene.

\subsection{Evaluation Setup}
\label{sect:exp_setup}
\paragraph{Scene Reconstruction Dataset.}
We reconstruct driving scenes from the Waymo Open Motion Dataset~\citep{ettinger2021large}.
We select 12 challenging sequences featuring multiple vehicles, complex driving maneuvers, and diverse lighting conditions.
Each scene is manually segmented into training and evaluation clips.
In our experiments, we simultaneously use videos from frontal and two side cameras to capture a wide field of view and track cars moving across the scene.
More evaluation setup details are provided in \app{eval_setup}.

\paragraph{Scene Extrapolation with Novel View Synthesis.}
For our evaluation, we selected driving scenes involving U-turns, intersection crossings, and parking departures — common accident scenarios that expose vehicles from diverse viewpoints and pose significant challenges for reconstruction.
Each sequence was manually divided into training and testing subsets at the midpoint of the maneuver.
We use the whole sequence to reconstruct the background and then remove the cars using the annotated bounding boxes in the Waymo dataset.
Car reconstruction relies solely on the training portion, while the remaining frames are reserved for evaluating scene reconstruction quality.

Our goal is to generate realistic novel views by extrapolating beyond the observed data.
Specifically, we insert the reconstructed car models into the background according to location and orientation specified by the bounding boxes on the holdout sequence.
This setup intentionally tests the model under configurations that differ from the training views, while ensuring no test data leaks into the reconstruction process.

\paragraph{Baselines.} 
We compare \ourmethod with the scene reconstruction Gaussian splatting-based methods that were previously considered for novel view synthesis: Street-Gaussians (SG)~\citep{yan2024street}, AutoSplat~\citep{khan2024autosplat} (our implementation), and HUGS~\citep{zhou2024hugs}.
Details on training and evaluation of baselines are given in \app{baseline_details}.


\subsection{Main Experiments}
\label{sect:exp_main_results}
\paragraph{Qualitative Evaluation.}
%
First, we provide visual scene reconstruction results for qualitative analysis. 
In~\fig{qualitative_frame_comparison}, we compare renderings on the training and hold-out frames.
While SG, AutoSplat, and HUGS reproduce the training frames with high accuracy, their reconstructions tend to break down under novel viewpoints, causing vehicles to fall apart or distort.
In contrast, our method, though slightly less precise on training frames, demonstrates substantially greater robustness to unseen configurations.
Additional visual examples are shown in Figures~\ref{fig:qualitative_frame_comparison_v2},~\ref{fig:qualitative_frame_comparison_v3}.
We also provide the visualizations with modified trajectories in~\fig{results_grid}.

\begin{table}[h]
\small
\centering
\vspace{-1mm}
\caption{Comparison in terms of tracking and segmentation metrics.}
\label{tab:main_metrics}
\resizebox{1.0\linewidth}{!}{
\begin{tabular}{l|cccc}
\toprule
Model & MOTA $\uparrow$ & MOTP $\downarrow$ & IDF1 $\uparrow$ &  Segmentation IoU  $\uparrow$ \\
\midrule
Street-GS~\citep{yan2024street} & 0.654 & \bf{0.105} & 0.776 & 0.556  \\
HUGS~\citep{zhou2024hugs} & 0.556 & 0.221  & 0.699 & 0.333  \\
AutoSplat\textsuperscript{*}~\citep{khan2024autosplat} & 0.589 & 0.154 & 0.716 & 0.489  \\
\midrule
\ourmethod (Ours) & \bf{0.841} & 0.138  & \bf{0.913} & \bf{0.818}  \\
\midrule
\multicolumn{5}{l}{\textsuperscript{*}Denotes our reimplementation.} 
\end{tabular}
}
\vspace{-7mm}
\end{table}


\begin{figure*}[!t]
    \centering
    \vspace{-5mm}
    \includegraphics[width=0.95\linewidth]{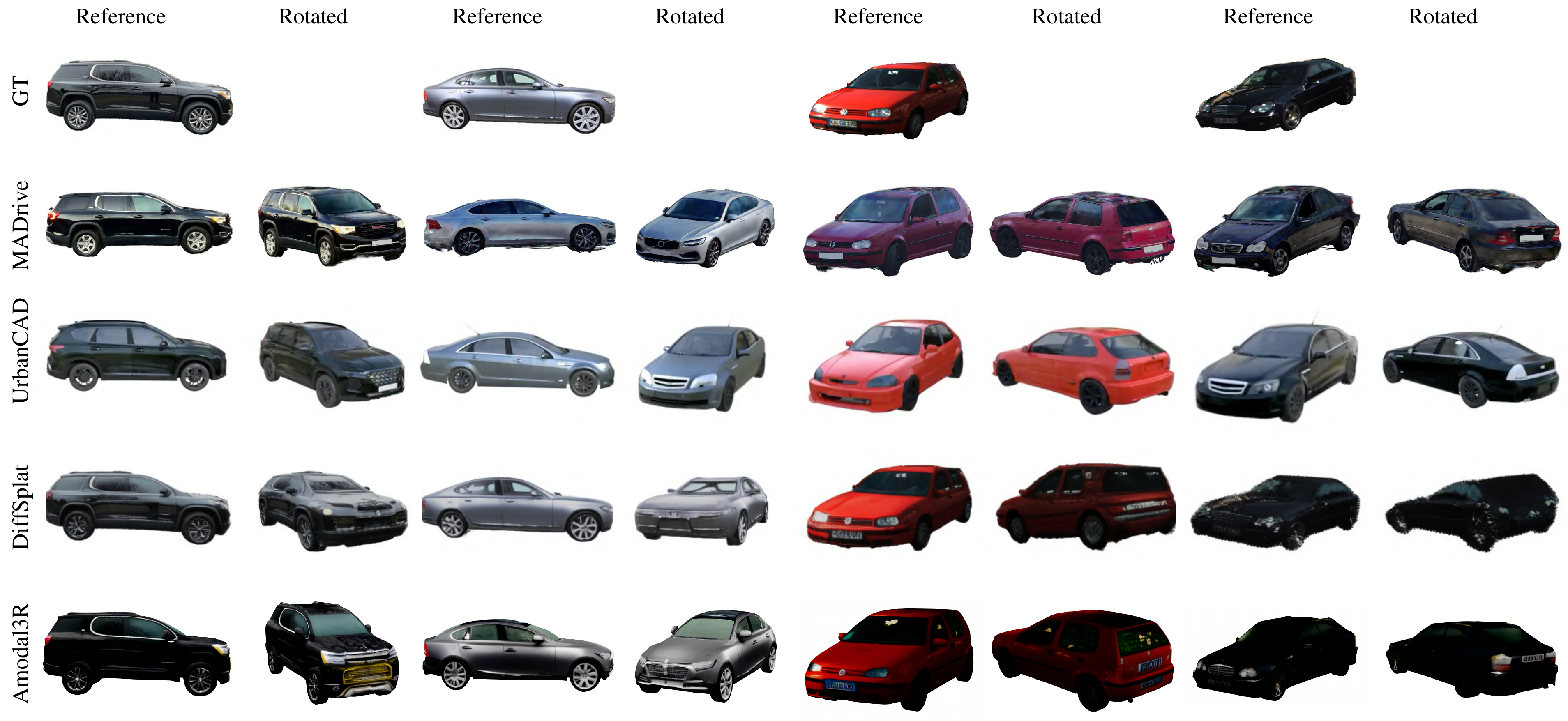}
    \vspace{-2mm}
    \caption{
\textbf{Qualitative comparison of reconstructed vehicle assets} based on queries from KITTI-360 (reference and rotated viewpoints).
\ourmethod, trained on the \ourdataset dataset, 
produces 3D assets that better preserve shape consistency and visual realism across views compared to CAD-based and image-to-3D generative baselines.
}
    \label{fig:qualitative_car_comparison}
    \vspace{-4mm}
\end{figure*}

\paragraph{Quantitative Evaluation.}
In our main experiments, we evaluate tracking and segmentation performance on synthesized \textbf{test} frames.
Specifically, we apply state-of-the-art tracking and segmentation models to both synthesized and ground truth frames and compare their outputs using established metrics for each task.
For tracking, we use BotSort~\citep{aharon2022bot} with a YOLOv8n backbone and report Multiple Object Tracking Accuracy (MOTA$\uparrow$), Multiple Object Tracking Precision (MOTP$\downarrow$), and the identity F1 score (IDF1)~\citep{milan2016mot16}.
For segmentation, we compute the average intersection-over-union (IoU) using instance segmentation masks predicted by Mask2Former~\citep{cheng2022maskedattention}.

\tab{main_metrics} compares \ourmethod against baseline approaches.
\ourmethod achieves substantially higher scores in two tracking metrics (MOTA and IDF1) and in the segmentation metric (IoU), confirming improved scene consistency and object reconstruction quality.
This observation is also supported by the visual examples provided in~\fig{qualitative_frame_comparison}. 

The lower MOTP score of \ourmethod compared to Street-GS arises from Street-GS’s better initial alignment in early test frames; however, later frames—where Street-GS tracking often fails—are excluded from the MOTP calculation.
Per-scene results for all 12 sequences are provided in \app{detailed_results}, and the choice of reference masks used in the evaluation protocol is discussed in \app{choice_reference_evaluation}.

\subsection{Ablation Study}
\label{sect:ablation_study}
\paragraph{Retrieval.}


Here, we evaluate the retrieval module in isolation to assess how accurately the retrieved cars correspond to the original vehicles in the scene.

We compare retrieval performance on \ourdataset against 3DRealCars~\citep{du20243drealcar}, a high-quality publicly available dataset containing 2,500 car assets.
To evaluate retrieval quality, we first compute the average L2 distance between each car image from the driving scene and its nearest neighbor in the memory bank, using SigLIP2~\citep{tschannen2025siglip} as the image feature extractor.
Then, we provide accuracy obtained with the Qwen2.5-VL-32B-Instruct model, which compares cars based on brand, model, color, and car type.
For a fair comparison, we do not use the color filtering in this experiment.

\tab{dataset_info} reports retrieval accuracy across different attributes, along with the average L2 distance to the closest instance.
Cars retrieved from \ourdataset more closely match the driving scene vehicles, which we attribute to the dataset’s larger scale and diversity.

Notably, \tab{dataset_info} highlights that retrieval based solely on feature embeddings often ignores car color, despite its importance for accurate vehicle replacement.
A similar limitation has been observed with non-vision-language encoders such as DINOv2~\citep{oquab2023dinov2}.
Additional results in~\app{qualitative_retrieval_comparison} show that applying a color-based pre-filtering improves color consistency between the retrieved and target vehicles.

\begin{table}[h]
    \centering
    \caption{
        Retrieval performance w/o color filtering in terms of accuracy on the car brand, model, color and type and the distance to the closest instance for the \ourdataset and 3DRealCar~\citep{du20243drealcar} datasets.
        \ourdataset enables more accurate retrieval of cars across all attributes.
    }
    \label{tab:dataset_info}
    \resizebox{\linewidth}{!}{%
        \begin{tabular}{lllll|l}
    \toprule
    \textbf{Dataset} & \textbf{Brand} $\uparrow$ & \textbf{Model} $\uparrow$ & \textbf{Color} $\uparrow$ & \textbf{Car Type} $\uparrow$ & \textbf{Distance} $\downarrow$ \\
    \midrule
    3DRealCars & 0.626 & 0.503 & 0.508 & 0.888 & 0.502\\
    \ourdataset        & 0.750 & 0.663 & 0.533 & 0.913 & 0.445 \\
    \bottomrule
    \end{tabular}%
    }
    \vspace{-4mm}
\end{table}




\begin{figure*}[!t]
    \centering
    \vspace{-3mm}
    \includegraphics[width=0.98\linewidth]{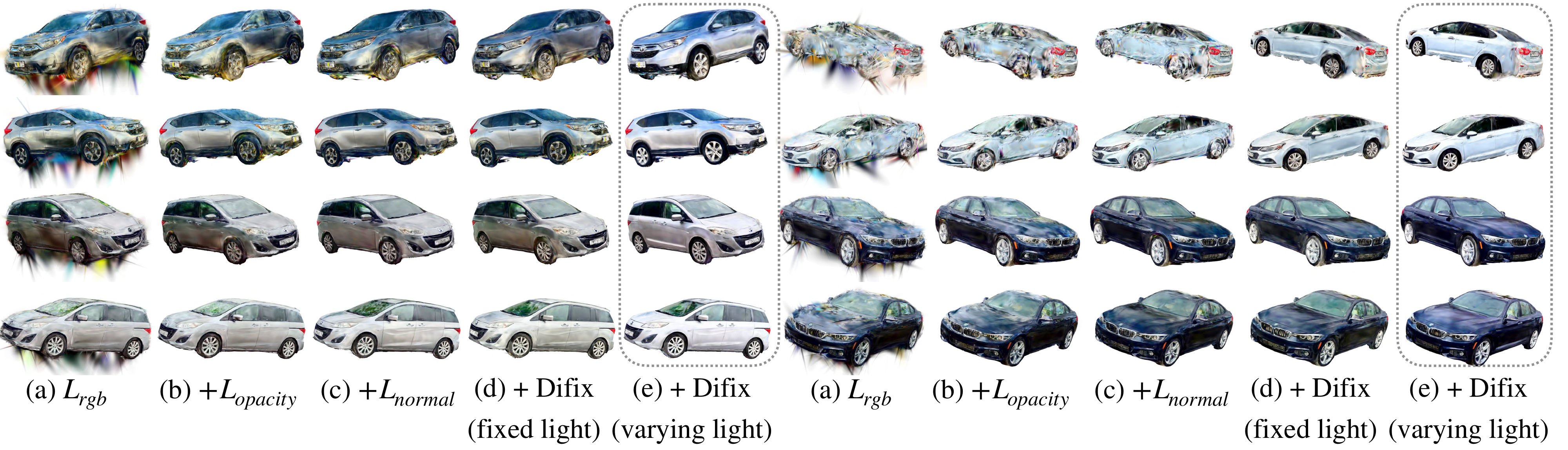}
    \vspace{-2mm}
    \caption{
Qualitative ablation of reconstruction components, progressively adding each regularizer to the previous configuration.
Starting without regularization (a), the reconstruction shows shape and texture artifacts with uneven edges.
Adding opacity regularization (b) improves edge quality, while normal regularization (c) enhances surface smoothness.
Incorporating synthetic data (d) further refines the results, and training with synthetic frames under varying lighting (e) helps disentangle illumination from object color, producing cleaner albedo and overall more consistent reconstructions.
}
    \label{fig:reconstruction_ablation}
    \vspace{-4mm}
\end{figure*}

\paragraph{Car reconstruction.}
We provide a qualitative comparison with other car reconstruction approaches in \fig{qualitative_car_comparison}, where we visualized 
reconstruction alternatives. 
Given a query frame from the KITTI dataset, we compared the proposed approach with three alternatives:
matching with a car model from a CAD dataset (UrbanCAD, \citep{lu2024urbancad}), and running a cutting edge image-to-3D models (DiffSplat \citep{lin2025diffsplat}, Amodal3R \citep{wu2025amodal3r}).
Even though the latter closely matches the query frame, the second view indicates a subpar geometry recovery.
Compared to other methods, we see that the diversity of our dataset allows \ourmethod to obtain models that closely match the query frame in terms of appearance (e.g., color, shape) and realism.

We also quantitatively compare the realism of reconstructed cars for the retrieval-based approach of \ourmethod and the recent image-to-3D generative model, Amodal3R. 


For this, we collected random crops using a hold-out set of real cars from the Waymo dataset and reconstructed car models with \ourmethod and Amodal3R. Then we rendered reconstructed 3D models from 360 degrees. 
In Table \ref{tab:car_fid}, we provide FID \citep{heusel2017gans} and KID \citep{bińkowski2018demystifying} between the set of renderings for the reconstructed models with \ourmethod and Amodal3R and the set of real images of hold-out cars from \ourdataset. The results support our claim that the retrieval-augmented approach yields cars with better realism. We provide the details in \app{car_fid}.

\begin{table}[h]
\centering
\caption{
Quantitative evaluation of the realism for reconstructed car models.
}\label{tab:car_fid}
\resizebox{0.7\linewidth}{!}{%
\begin{tabular}{l|c c}
\hline
               & FID\ $\downarrow$ & KID$\times 10^3$\ $\downarrow$ \\ \hline
Amodal3R       & 81.65                            & 51.91                                                      \\
\ourmethod & 62.64                            & 39.40                                                      \\ \hline
\end{tabular}
}
\vspace{0mm}
\end{table}

We further ablate the components of the proposed reconstruction algorithm.
\fig{reconstruction_ablation} shows the recovered geometry and albedo for several cars.
Reconstruction without regularization (a) produces noticeable artifacts in both shape and texture, leading to uneven edges after background removal.
Introducing opacity regularization to suppress the background (b) improves edge quality.
Adding normal regularization (c) further enhances surface smoothness and consistency.
Incorporating synthetic views enhanced with Difix (d) improves reconstruction quality in some cases.
Finally, training with synthetic frames rendered under varying lighting conditions (e) helps disentangle scene illumination from object color, resulting in more uniform albedo and overall cleaner reconstructions.



\begin{figure}[ht]
  \vspace{-1mm}
  \begin{center}
  \begin{minipage}{0.45\textwidth}
    \includegraphics[width=\linewidth]{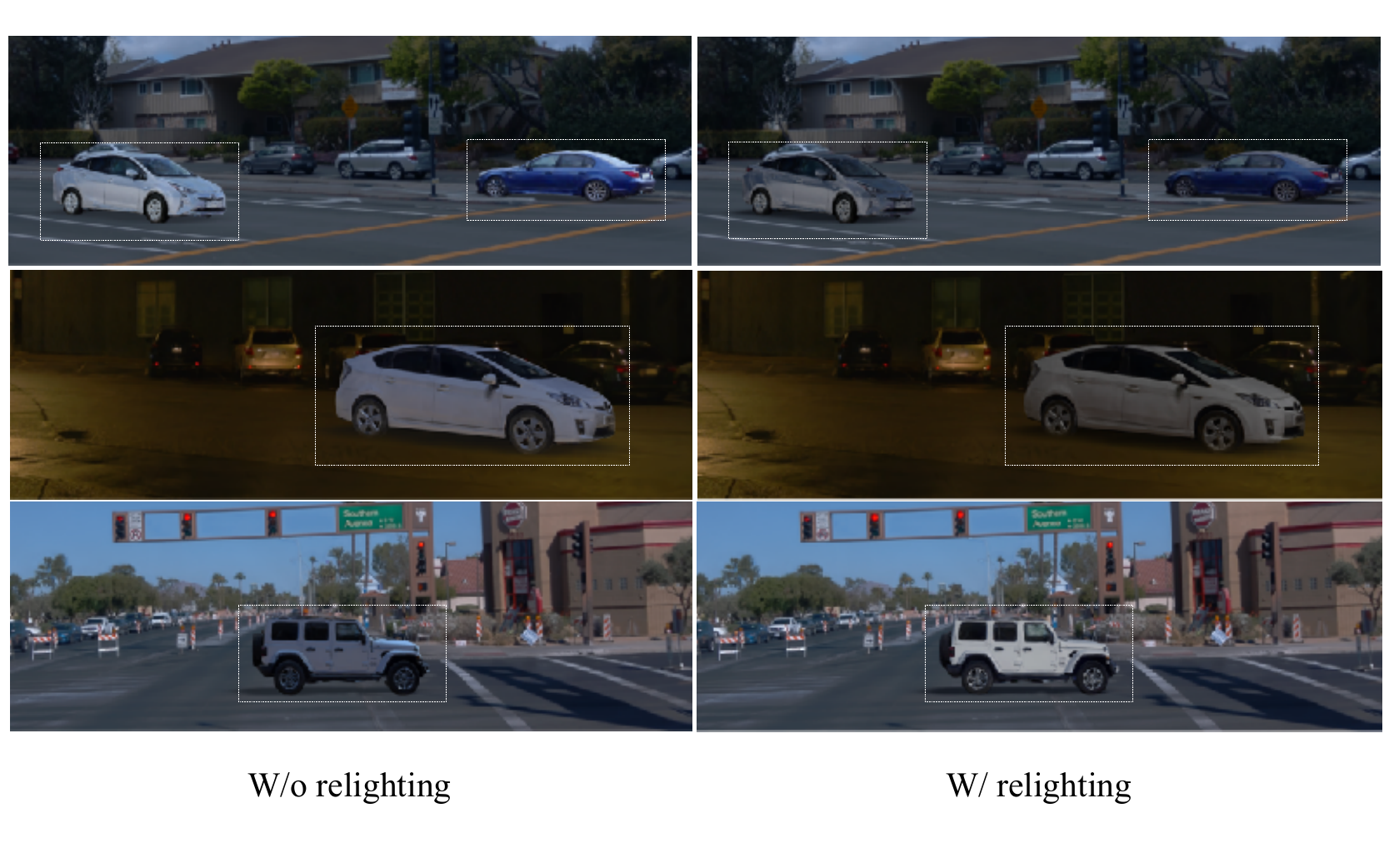}
  \end{minipage}
  \end{center}
  \vspace{-9mm}
  \caption{\textbf{Relighting ablation.} Rendered hold-out frames without ({\color{analysisblue} Left}) and with ({\color{analysispurple} Right}) relighting.}
  \vspace{0mm}
  \label{fig:qualitative_relighting_comparison}
\end{figure}

\paragraph{Relighting.}\label{sect:exp_relighting}
We performed a qualitative comparison to evaluate the impact of the proposed relighting scheme.
For several scenes, we reconstructed frames both with and without the relighting module.
As shown in~\fig{qualitative_relighting_comparison}, the relighting module effectively adjusts model colors to the surrounding illumination, reducing visual inconsistencies and making the inserted vehicles appear more naturally integrated into the scene.

\vspace{-4mm}
\section{Conclusion}
This work presents \ourmethod, a novel driving scene reconstruction approach specifically designed to model significantly altered vehicle positions.
Powered by \ourdataset, our large-scale multi-view car dataset, \ourmethod replaces dynamic vehicles in a scene with similar car instances from the database.
We believe that \ourmethod could make a step towards modeling multiple potential outcomes for analyzing an autonomous driving system's behavior in safety-critical situations.

Although our generated future frames look promising, they still differ from the ground truth, as discussed in~\app{limitation}.
At the moment, we only use the future frame to create the retrieval query, but it could also help adapt the inserted asset to better match the observed vehicle in the future work.
Another direction is to include more advanced relighting methods that better capture reflections and light interactions, making the results more realistic under new lighting conditions.
\newpage

{
    \small
    \bibliographystyle{ieeenat_fullname}
    \bibliography{sample-base}
}

\appendix
\clearpage
\section*{Supplementary material}
\section{Urban Scene Reconstruction with NeRFs}
\label{appendix:related_work_nerf}
NeRFs \citep{mildenhall2020nerf} can be used to model dynamic urban scenes. SUDS \citep{turki2023suds} uses a single network for dynamic actors, which limits the possibility of altering the behavior of the actors. EmerNeRF \citep{yang2024emernerf} follows a similar idea to SUDS by decomposing the scene purely into static and dynamic components. NeuRAD \citep{tonderski2024neurad} takes advantage of monocular or LiDAR-based 3D bounding box predictions and proposes a joint optimization of object poses during the reconstruction process. Although these methods produce reasonable results, they are still 1) limited to the high training cost and low rendering speed; or 2) do not address extrapolation of future vehicle appearance far beyond the original camera views.

\section{Data Collection Details}
\label{appendix:data_collection}

The initial database contained ${\sim}95,000$ car videos of ${\sim}100$ views on average.
The first filtering stage includes the filtering of low quality and overly dark images with the CLIP-IQA model~\citep{wang2022exploring}, discarding frames with a score $<0.2$.
Then, we use Qwen-2.5-VL-Instruct (7B)~\citep{yang2024qwen2} to respond several questions for each frame:

\begin{itemize}
    \item ``Does the image depict a car?''
    \item ``Is the car directly occluded?''
    \item ``Does the image depict the car interior?''
    \item ``Does a hand or finger block the view?''
    \item ``Is the car door open?''
    \item ``Does the image mainly depict the car window?''
\end{itemize}

Based on the responses, we filter out the corresponding frames or, in some cases, entire car instances.
Also, if fewer than $45$ valid frames remain for a given instance, the entire instance is discarded.

\section{Evaluation Setup Details}
\label{appendix:eval_setup}

For scene reconstruction evaluation, we selected 12 scenes from the Waymo Open Dataset~\citep{sun2020scalability}, with labels listed in \tab{scenes_names}. 
This table also provides the correspondence between the original scene labels from the \href{https://console.cloud.google.com/storage/browser/waymo_open_dataset_v_1_4_3}{Waymo Cloud Storage} and the short names used in our work.
We split each scene into training and testing subsets based on time (\tab{train_test_split_time}) and camera selection (\tab{train_test_split_cameras}).
Specifically, frames with indices $i^{\text{train}}$, where $i^{\text{train}} \in [i_{\text{start}}^{\text{train}}, i_{\text{end}}^{\text{train}}]$, were used for training.
For evaluation, we used frames $i^{\text{test}} \in [i_{\text{start}}^{\text{test}}, i_{\text{end}}^{\text{test}}]$, with all split indices provided in \tab{train_test_split_time}.

\begin{table}[htb]
\small
\centering
\caption{Waymo scenes used for evaluation of scene reconstruction.}
\label{tab:scenes_names}
\resizebox{1.0\linewidth}{!}{
\begin{tabular}{c|c}
\toprule
Label & Scene name \\
\midrule
1231623110026745648\_480\_000\_500\_000 & 123 \\
1432918953215186312\_5101\_320\_5121\_320 & 143 \\
1906113358876584689\_1359\_560\_1379\_560 & 190 \\
10500357041547037089\_1474\_800\_1494\_800 & 105 \\
10940952441434390507\_1888\_710\_1908\_710 & 109 \\
16504318334867223853\_480\_000\_500\_000 & 165 \\
17407069523496279950\_4354\_900\_4374\_900 & 174 \\
18025338595059503802\_571\_216\_591\_216 & 180 \\
14183710428479823719\_3140\_000\_3160\_000 & 141 \\
15834329472172048691\_2956\_760\_2976\_760 & 158 \\
17647858901077503501\_1500\_000\_1520\_000 & 176 \\
7799671367768576481\_260\_000\_280\_000 & 779 \\
\bottomrule
\end{tabular}
}
\end{table}

\begin{table}[htb]
\small
\centering
\caption{Train and test frame splits for Waymo scenes over time.
All values, except those in the leftmost column, indicate frame indices starting from 0.}
\label{tab:train_test_split_time}
\begin{tabular}{c|cccccc}
\toprule
Scene name &  $i_{start}^{train}$ &  $i_{end}^{train}$ &  $i_{start}^{test}$ & $i_{end}^{test}$ \\
\midrule
123 & 106 & 116 & 117 & 175 \\
143 & 43 & 53 & 54 & 62 \\
190 & 115 & 125 & 126 & 137 \\
105 & 164 & 174 & 175 & 196 \\
109 & 1 & 16 & 17 & 55 \\
165 & 7 & 40 & 41 & 111 \\
174 & 34 & 51 & 52 & 72 \\
180 & 49 & 55 & 56 & 68 \\
141 & 60 & 80 & 81 & 117 \\
158 & 44 & 62 & 63 & 100 \\
176 & 31 & 42 & 43 & 67 \\
779 & 50 & 65 & 66 & 84 \\
\bottomrule
\end{tabular}
\end{table}

\begin{table}[htb]
\small
\centering
\caption{Train and test frame splits for Waymo scenes based on camera selection.}
\label{tab:train_test_split_cameras}
\resizebox{1.0\linewidth}{!}{
\begin{tabular}{c|cc}
\toprule
Scene name & Train cameras & Test cameras \\
\midrule
123 & frontal, frontal left & frontal, frontal left \\
143 & frontal, frontal left & frontal, frontal left \\
190 & frontal, frontal left & frontal, frontal left \\
105 & frontal, frontal left & frontal \\
109 & frontal, frontal right & frontal right \\
165 & frontal, frontal left & frontal, frontal left \\
174 & frontal & frontal \\
180 & frontal, frontal right & frontal, frontal right \\
141 & frontal & frontal \\
158 & frontal & frontal \\
176 & frontal & frontal \\
779 & frontal, frontal left, frontal right & frontal, frontal right \\
\bottomrule
\end{tabular}
}
\end{table}

\section{Per-Scene Quantitative Evaluation.}
\label{appendix:detailed_results}
In addition to the aggregated results in \tab{main_metrics}, we report per-scene metric values in \tab{test_mota_scenes}, \tab{test_motp_scenes}, \tab{test_idf1_scenes}, and \tab{test_iou_scenes}, corresponding to MOTA, MOTP, IDF1, and IoU, respectively.
We observe that \ourmethod consistently outperforms the baselines across most scenes.

\begin{table}[ht]
\small
\centering
\caption{Mean MOTA $\uparrow$ results on test frames for all Waymo scenes.}
\label{tab:test_mota_scenes}
\begin{tabular}{c|ccc|c}
\toprule
Scene name & SG & HUGS & AutoSplat & \ourmethod \\
\midrule
123 & 0.687 & 0.685 & 0.327 & 0.887 \\
143 & 0.650 & 0.513 & 0.600 & 0.825 \\
190 & 0.787 & 0.795 & 0.904 & 0.858 \\
105 & 0.906 & 0.656 & 0.742 & 0.906 \\
109 & 0.242 & 0.448 & 0.605 & 0.925 \\
165 & 0.684 & 0.461 & 0.788 & 0.883 \\
174 & 0.809 & 0.886 & 0.830 & 0.936 \\
180 & 0.611 & 0.528 & 0.695 & 0.778 \\
141 & 0.667 & 0.607 & 0.163 & 0.767 \\
158 & 0.423 & 0.233 & 0.681 & 0.639 \\
176 & 0.727 & 0.562 & 0.176 & 0.912 \\
779 & 0.661 & 0.296 & 0.545 & 0.779 \\
\bottomrule
\end{tabular}
\end{table}

\begin{table}[ht]
\small
\centering
\caption{Mean MOTP $\downarrow$ results on test frames for all Waymo scenes.}
\label{tab:test_motp_scenes}
\begin{tabular}{c|ccc|c}
\toprule
Scene name & SG & HUGS & AutoSplat & \ourmethod \\
\midrule
123 & 0.073 & 0.093 & 0.099 & 0.079 \\
143 & 0.114 & 0.461 & 0.095 & 0.203 \\
190 & 0.088 & 0.112 & 0.115 & 0.144 \\
105 & 0.073 & 0.262 & 0.222 & 0.118 \\
109 & 0.093 & 0.132 & 0.094 & 0.122 \\
165 & 0.125 & 0.202 & 0.119 & 0.149 \\
174 & 0.075 & 0.886 & 0.078 & 0.093 \\
180 & 0.150 & 0.231 & 0.194 & 0.195 \\
141 & 0.119 & 0.261 & 0.237 & 0.179 \\
158 & 0.087 & 0.128 & 0.123 & 0.119 \\
176 & 0.093 & 0.246 & 0.167 & 0.072 \\
779 & 0.176 & 0.443 & 0.305 & 0.180 \\
\bottomrule
\end{tabular}
\end{table}

\begin{table}[ht]
\small
\centering
\caption{Mean IDF1 $\uparrow$ results on test frames for all Waymo scenes.}
\label{tab:test_idf1_scenes}
\begin{tabular}{c|ccc|c}
\toprule
Scene name & SG & HUGS & AutoSplat & \ourmethod \\
\midrule
123 & 0.804 & 0.806 & 0.475 & 0.940 \\
143 & 0.787 & 0.709 & 0.750 & 0.904 \\
190 & 0.880 & 0.887 & 0.950 & 0.924 \\
105 & 0.952 & 0.780 & 0.877 & 0.951 \\
109 & 0.390 & 0.619 & 0.754 & 0.961 \\
165 & 0.806 & 0.612 & 0.894 & 0.936 \\
174 & 0.894 & 0.940 & 0.907 & 0.967 \\
180 & 0.753 & 0.709 & 0.820 & 0.871 \\
141 & 0.805 & 0.698 & 0.278 & 0.866 \\
158 & 0.605 & 0.377 & 0.829 & 0.797 \\
176 & 0.847 & 0.720 & 0.316 & 0.955 \\
779 & 0.793 & 0.532 & 0.739 & 0.881 \\
\bottomrule
\end{tabular}
\end{table}


\begin{table}[ht]
\small
\centering
\caption{Mean IoU $\uparrow$ results on test frames for all Waymo scenes.}
\label{tab:test_iou_scenes}
\begin{tabular}{c|ccc|c}
\toprule
Scene name & SG & HUGS & AutoSplat  & \ourmethod \\
\midrule
123 & 0.753 & 0.608 & 0.500 & 0.866 \\
143 & 0.485 & 0.243 & 0.510 & 0.779 \\
190 & 0.707 & 0.519 & 0.740 & 0.846 \\
105 & 0.671 & 0.425 & 0.439 & 0.731 \\
109 & 0.499 & 0.246 & 0.419 & 0.832 \\
165 & 0.633 & 0.459 & 0.647 & 0.730 \\
174 & 0.695 & 0.581 & 0.655 & 0.829 \\
180 & 0.475 & 0.238 & 0.582 & 0.814 \\
141 & 0.607 & 0.196 & 0.226 & 0.765 \\
158 & 0.404 & 0.135 & 0.498 & 0.862 \\
176 & 0.499 & 0.187 & 0.263 & 0.886 \\
779 & 0.247 & 0.153 & 0.273 & 0.874 \\
\bottomrule
\end{tabular}
\end{table}

\section{Baseline Details}
\label{appendix:baseline_details}

\paragraph{Baselines training and evaluation.} We trained all baselines (Street-Gaussians, HUGS, and AutoSplat) for $10K$ iterations using the training frames with indices $i \in [i_{\text{start}}^{\text{train}}, i_{\text{end}}^{\text{train}}]$ as specified in \tab{train_test_split_time}.
Additionally, we trained the background models for both the baselines and \ourmethod for $30K$ iterations using all available frames.
These pretrained background models were then used during the rendering of the test frames ($i \in [i_{\text{start}}^{\text{test}}, i_{\text{end}}^{\text{test}}]$), on which we compute the metrics reported in \tab{test_mota_scenes}, \tab{test_motp_scenes}, \tab{test_idf1_scenes}, and \tab{test_iou_scenes}.

\paragraph{Street-Gaussians.} We used the official implementation available at \url{https://github.com/zju3dv/street_gaussians}.

\paragraph{HUGS.} We used the official implementation provided at \url{https://github.com/hyzhou404/HUGSIM}.

\paragraph{AutoSplat.} As no official implementation is publicly available, we re-implemented the core contributions of AutoSplat on top of the Street-Gaussians codebase.

\section{Choice of Reference Masks in the Evaluation}\label{appendix:choice_reference_evaluation}

In our validation setup, we used predictions from tracking and segmentation models on ground-truth images as targets, since the Waymo dataset lacks segmentation masks and 2D bounding boxes.

To evaluate whether cars in the synthesized frames are as identifiable as those in the original frames, we applied the same detection algorithm to both. Our method outperforms the baseline, primarily because our system inserts visually coherent cars on test frames by leveraging reconstructed models, whereas baseline approaches result in degraded or incomplete vehicle representations.

However, the inserted cars might be easier to detect. To test this, we conducted an additional experiment. Specifically, we generated a new set of detector targets by projecting the 3D bounding boxes provided in the Waymo dataset onto the image plane. We then evaluated the performance of the detector on both the ground-truth and synthesized (MADrive) frames using the new "ground-truth" annotation.

The results in \tab{gt_choice} show that the predictions on ground-truth images align slightly better with the projected 3D bounding boxes than those on the synthesized MADrive frames. This indicates that our inserted cars do not artificially simplify detection, supporting the validity of our evaluation.

\begin{table}[h]
\small
\centering
\caption{Comparison between detector performance on both the ground-truth and synthesized (MADrive) frames using projected 3D bounding boxes.}
\label{tab:gt_choice}
\begin{tabular}{l|ccc}
\toprule
Model & MOTA $\uparrow$ & MOTP $\downarrow$ & IDF1 $\uparrow$  \\
\midrule
GT frames & \textbf{0.879} & \textbf{0.270} & \textbf{0.928}   \\
\midrule
\ourmethod frames & 0.861 & 0.340  & 0.908 \\
\midrule
\end{tabular}
\end{table}

\section{Additional Retrieval Evaluation Results}
\label{appendix:qualitative_retrieval_comparison}
\begin{figure*}[t]
    \centering
    \vspace{0mm}
    \includegraphics[width=0.96\linewidth]{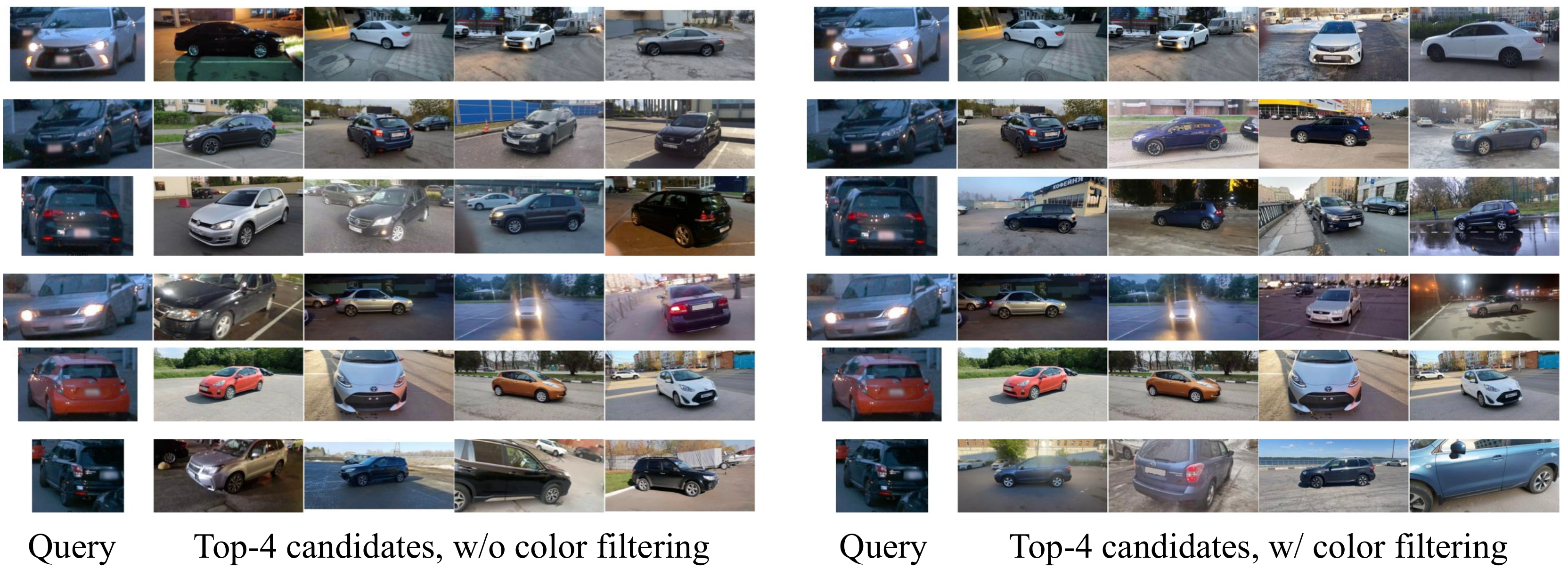}
    \vspace{-3mm}
    \caption{
        \textbf{Retrieval illustration.} Top-4 candidates retrieved using SigLIP 2 without ({\color{analysisblue} Left}) and with ({\color{analysispurple} Right}) color filtering.
    }
    \label{fig:qualitative_retrieval_comparison}
    \vspace{-2mm}
\end{figure*}

In this section, we provide an additional illustration of our retrieval algorithm.
As shown in our~\fig{qualitative_retrieval_comparison}, introducing a color-based pre-filter enhances the alignment of vehicle colors between retrieved candidates and the target.

\section{Evaluation of Realism for Reconstructed Models}
\label{appendix:car_fid}
To evaluate the realism of the reconstructed 3D car models using retrieval approach of \ourmethod and generative image-to-3D method of Amodal3R \citep{wu2025amodal3r}, we collected 38 random crops of hold-out cars from the Waymo dataset \citep{ettinger2021large}. We reconstructed 3D car models for all these crops with \ourmethod and Amodal3R and rendered them from 360 degrees with 60 renderings per reconstructed model. For Amodal3R, we also segment cars in crops using the YOLOv11 instance segmentation model \citep{yolo11_ultralytics}. For reference images, we took 30 hold-out cars from \ourdataset and collected near 2200 images for these cars. To reduce the domain shift, we excluded backgrounds from real images from \ourdataset with YOLOv11. Finally, we computed FID \citep{heusel2017gans} and KID \citep{bińkowski2018demystifying} scores between two sets: 1) the set of all renderings from the reconstructed 3D models with \ourmethod and Amodal3R, which we ran on 38 crops from Waymo; 2) the set of real images for 30 test cars from \ourdataset. Visually, we observed that the results of Amodal3R typically produce low-resolution cars with limited detail and a cartoon-like appearance, which explains the results in the Table \ref{tab:car_fid}. 


\section{Relighting Evaluation}

To assess the impact of our relighting module, we analyze the cars shown in Figure~\ref{fig:qualitative_relighting_comparison}.

Standard pixel-wise metrics (e.g., LPIPS) failed to capture meaningful differences, largely because geometric mismatches between the inserted asset and the original vehicle dominate these scores.
Instead, we evaluate relighting by comparing the color statistics of the integrated assets to those of the real cars.

We extract crops of the inserted vehicles and matching crops from the corresponding ground-truth frames.
Each crop is converted to the perceptually uniform CIELAB color space, and we compare their pixel-intensity distributions using the sliced Wasserstein distance~\cite{rubner2000earth}.
Average distances with standard deviations between different crops are provided in Table~\ref{tab:quantitative_relighting_comparison}.

Across all evaluated cars, relighting consistently reduces the discrepancy between synthetic and real crops.
A more detailed breakdown reveals that the improvement is especially pronounced in the lightness {\it(L*)} channel, while differences in the chromatic channels {\it(a*, b*)} are smaller—reflecting the fact that lighting primarily affects perceived brightness rather than intrinsic surface color.

\begin{table}[h]
\small
\centering
\begin{tabular}{c|ccc}
\toprule
               & $W_{1}$ & $W_{1, L^*}$         & $W_{1, (a^*, b^*)}$ \\
\midrule            
  \tiny{W/o relighting} & $6.74 \pm 1,83$         & $10.89 \pm 4.78$ & $3.82 \pm 3.34$ \\
  \tiny{W/ relighting}   & $\mathbf{3.15 \pm 0.90}$         & $\mathbf{3.54 \pm 2.75}$  & $2.89 \pm 1.30$ \\
\bottomrule
\end{tabular}
  \caption{Sliced Wasserstein distances between real and synthetic car crops in CIELAB space. Lower values indicate closer color distribution matches. Relighting consistently reduces the discrepancy, with the largest improvement observed in the lightness (L*) channel.}
\label{tab:quantitative_relighting_comparison}
\end{table}

\section{Additional Qualitative Comparisons and New Trajectories}
\begin{figure*}[t]
    \centering
        \includegraphics[width=\linewidth, page=6]{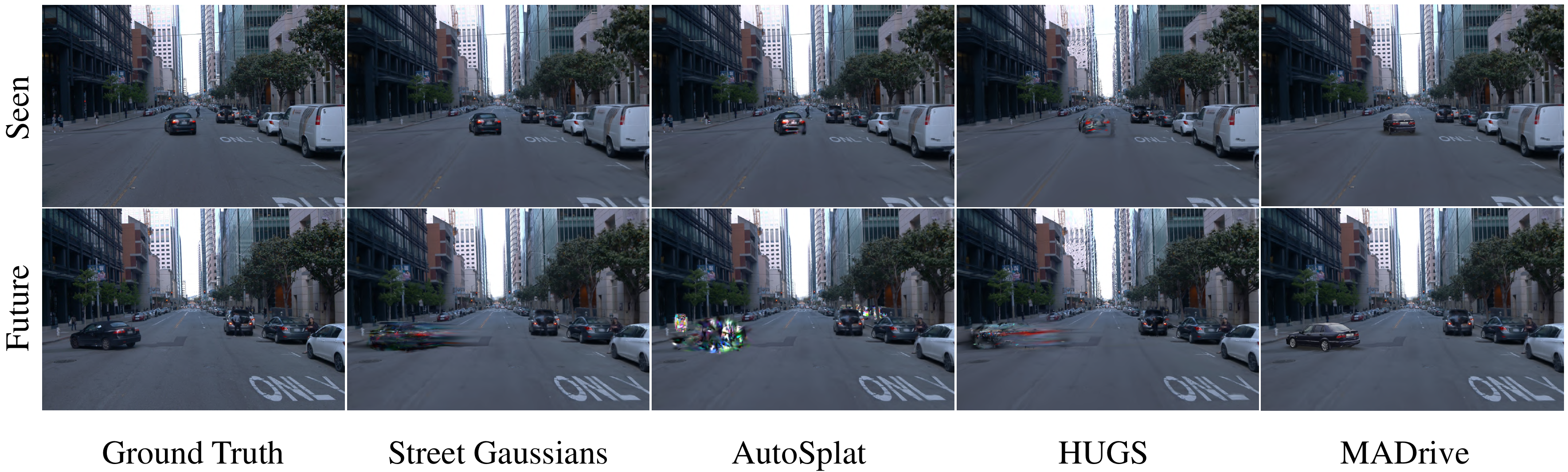}
        \includegraphics[width=\linewidth, page=5]{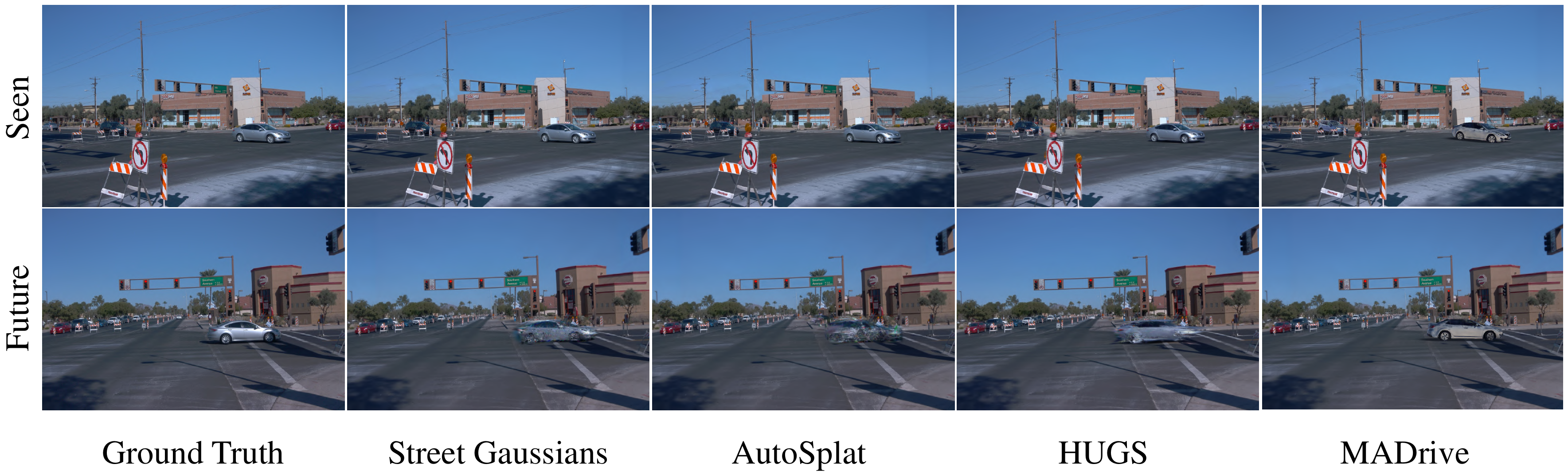}
        \includegraphics[width=\linewidth, page=2]{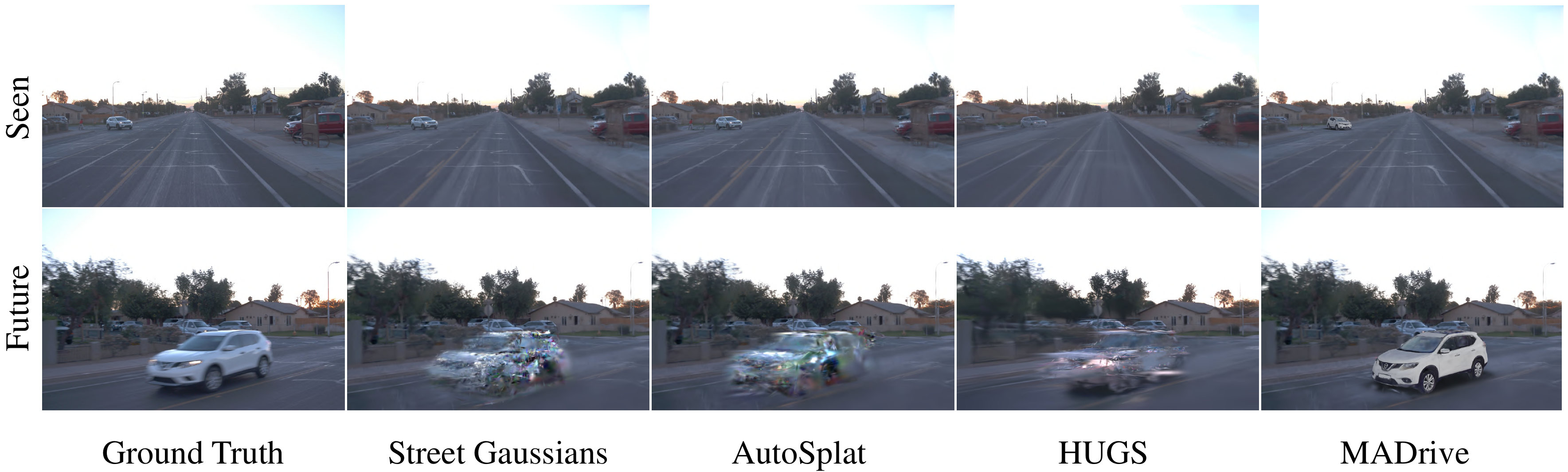}
        
    \caption{
        \textbf{Additional qualitative comparison} of \ourmethod with non-retrieval-based driving scene reconstruction methods.
        Reconstruction of the training views ({\color{analysisblue}Top}). 
        Reconstruction of the hold-out (future) views ({\color{analysisorange} Bottom}).
    }
    \label{fig:qualitative_frame_comparison_v2}
\end{figure*}

\begin{figure*}[t]
    \centering
        \includegraphics[width=\linewidth, page=3]{figures/qualitative_comparisons_all.pdf}
        \includegraphics[width=\linewidth, page=4]{figures/qualitative_comparisons_all.pdf}
        \includegraphics[width=\linewidth, page=1]{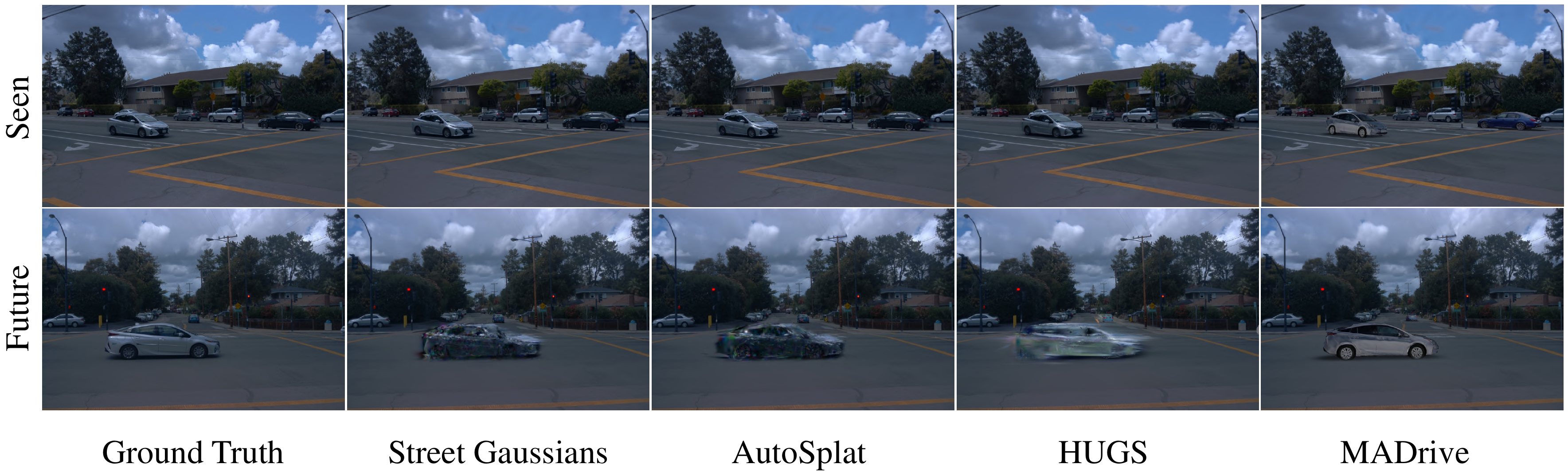}
        
    \caption{
        \textbf{Additional qualitative comparison} of \ourmethod with non-retrieval-based driving scene reconstruction methods.
        Reconstruction of the training views ({\color{analysisblue}Top}). 
        Reconstruction of the hold-out (future) views ({\color{analysisorange} Bottom}).
    }
    \label{fig:qualitative_frame_comparison_v3}
\end{figure*}

\begin{figure*}[htbp]
     \centering
     \begin{tabular}{rcccc}

         \rotatebox{90}{\parbox{2cm}{\centering\textbf{Original}}} &
         \includegraphics[width=0.2\linewidth]{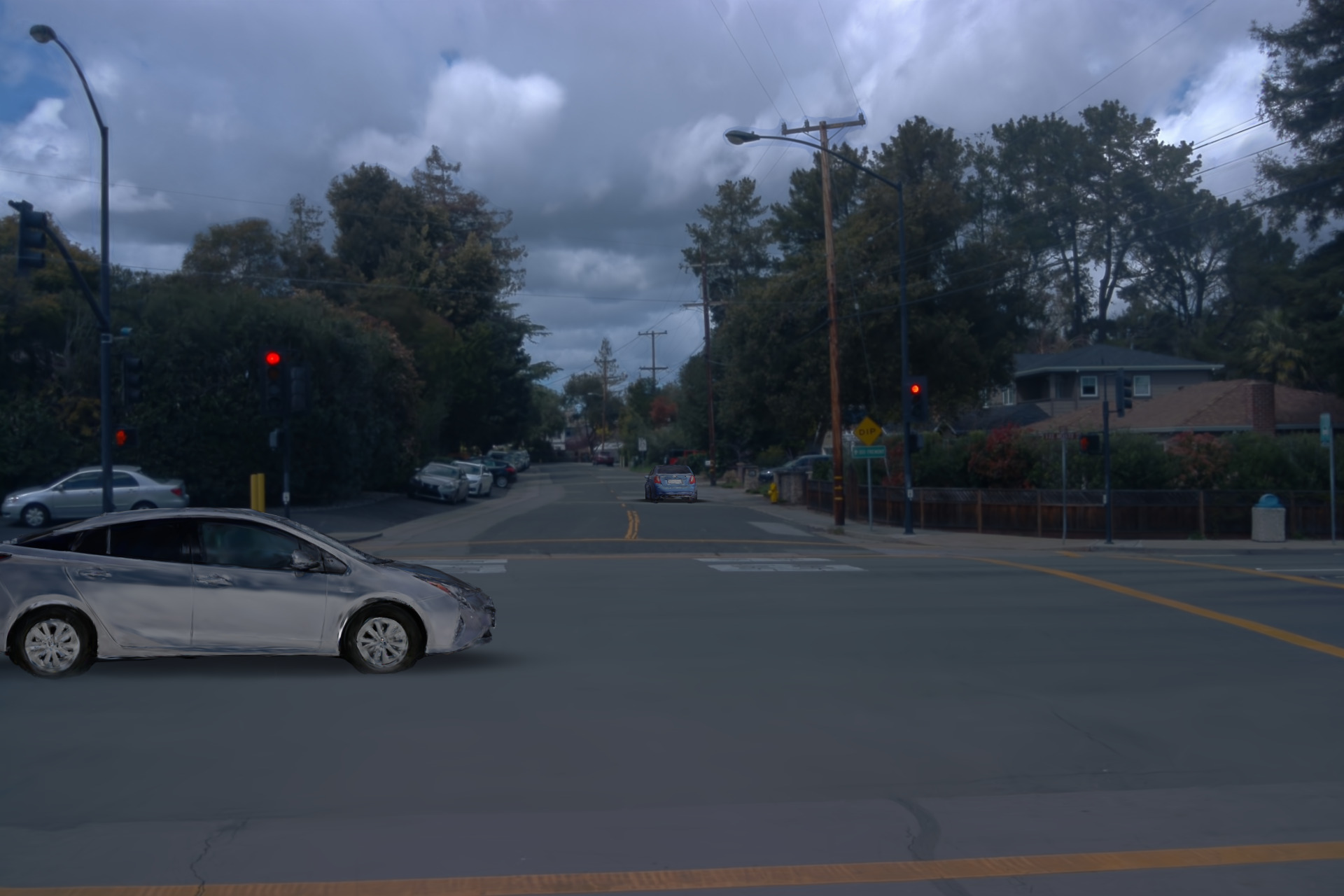} &
         \includegraphics[width=0.2\linewidth]{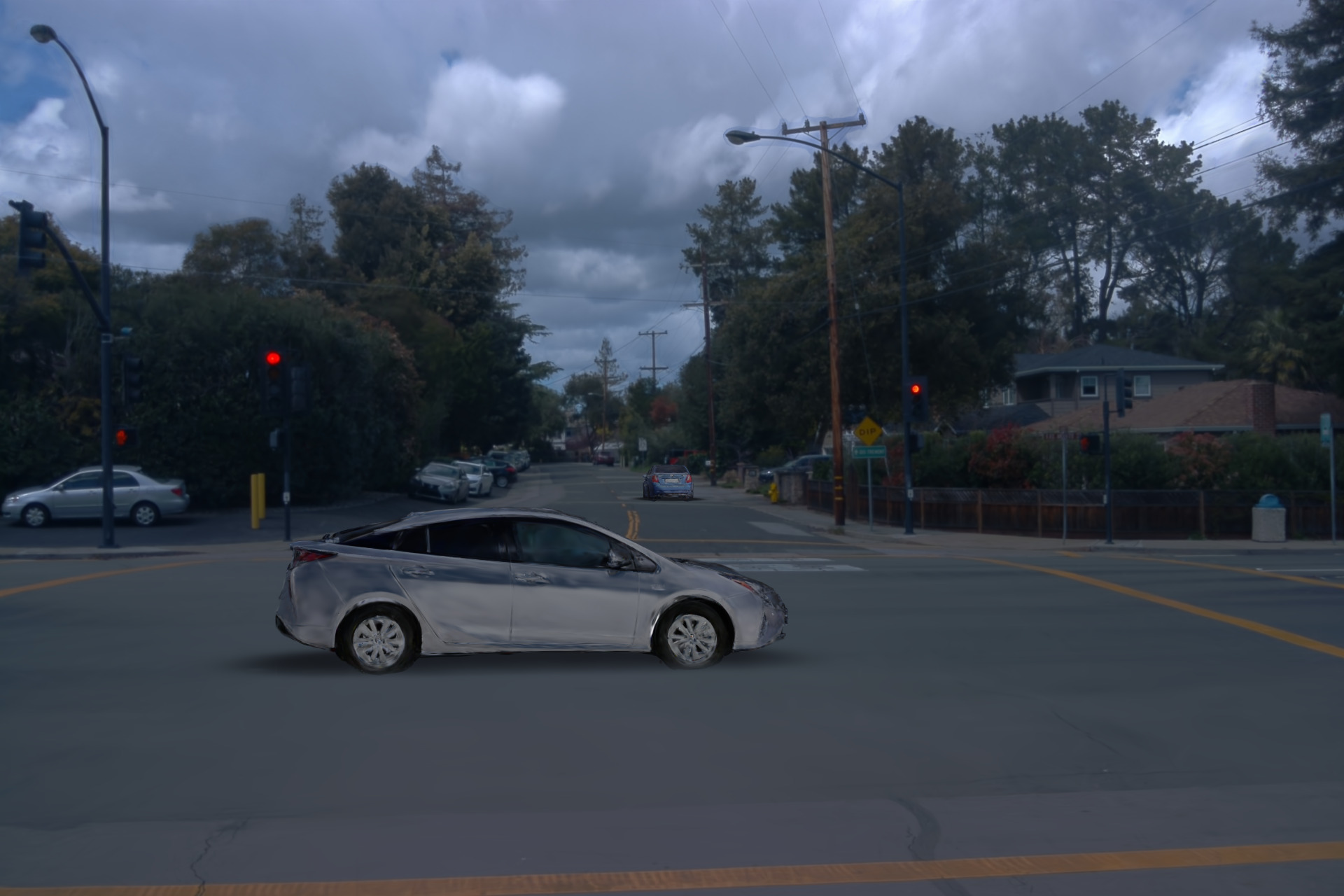} &
         \includegraphics[width=0.2\linewidth]{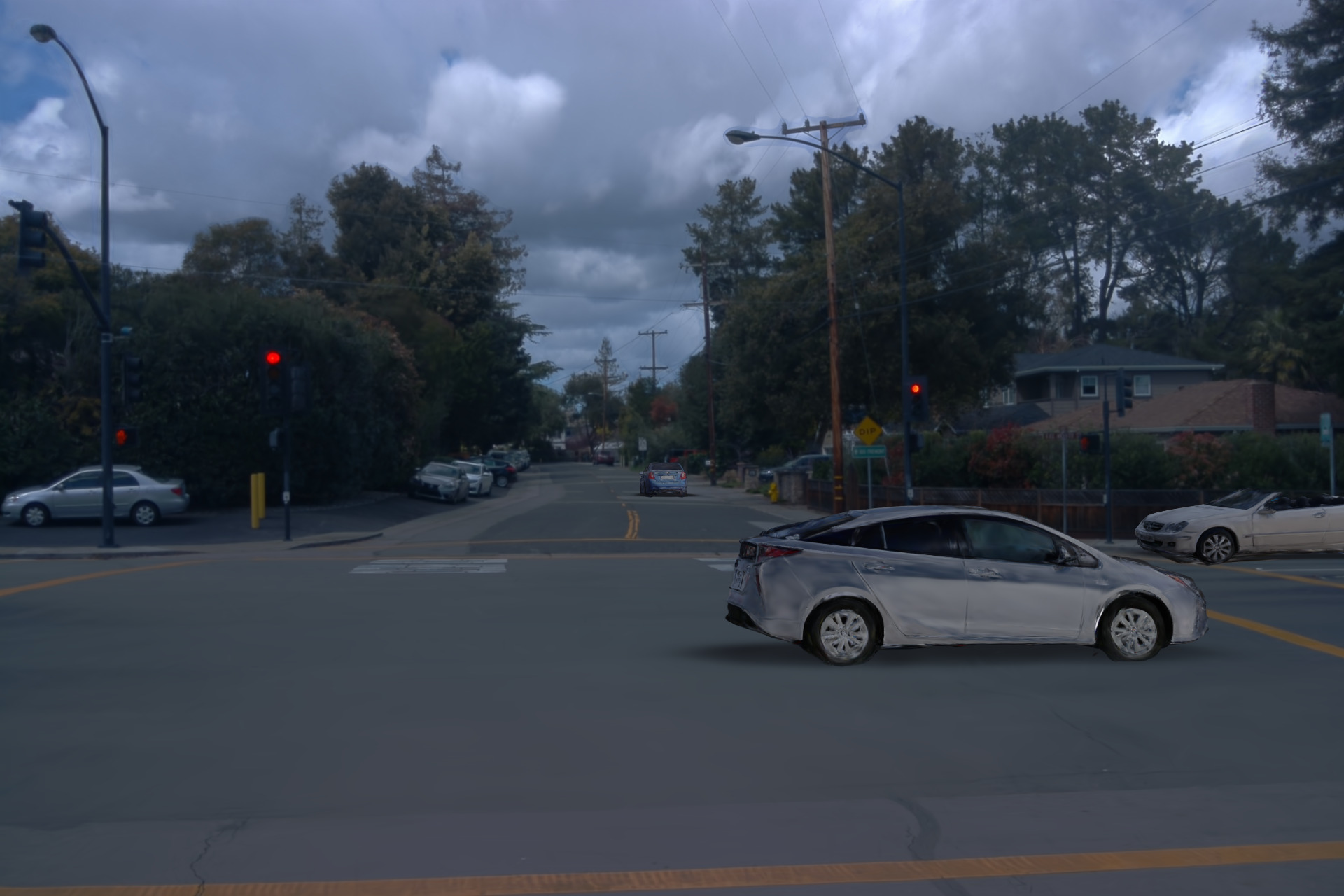} &
         \includegraphics[width=0.2\linewidth]{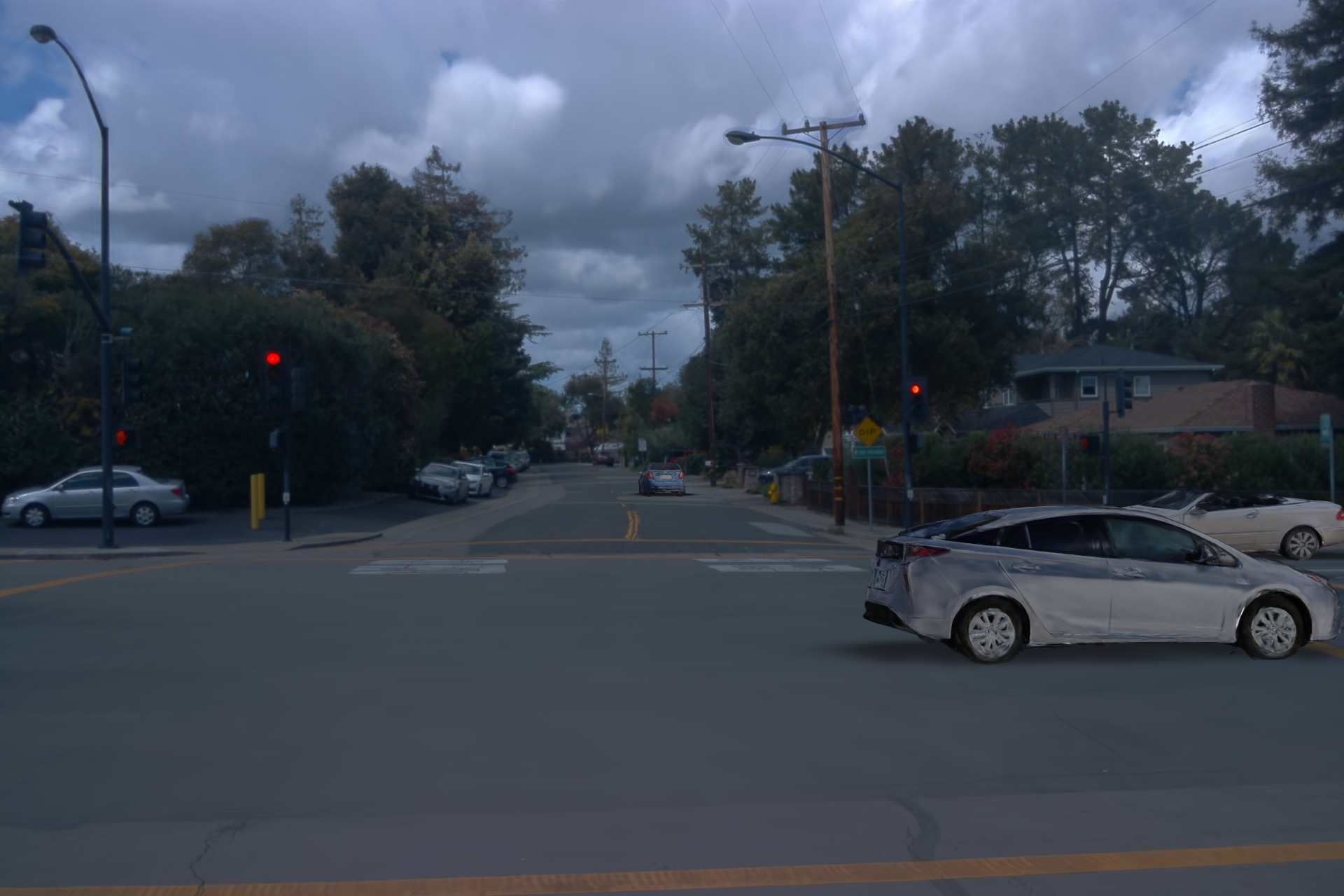} \\
         \rotatebox{90}{\parbox{2cm}{\centering\textbf{Modified}}} &
         \includegraphics[width=0.2\linewidth]{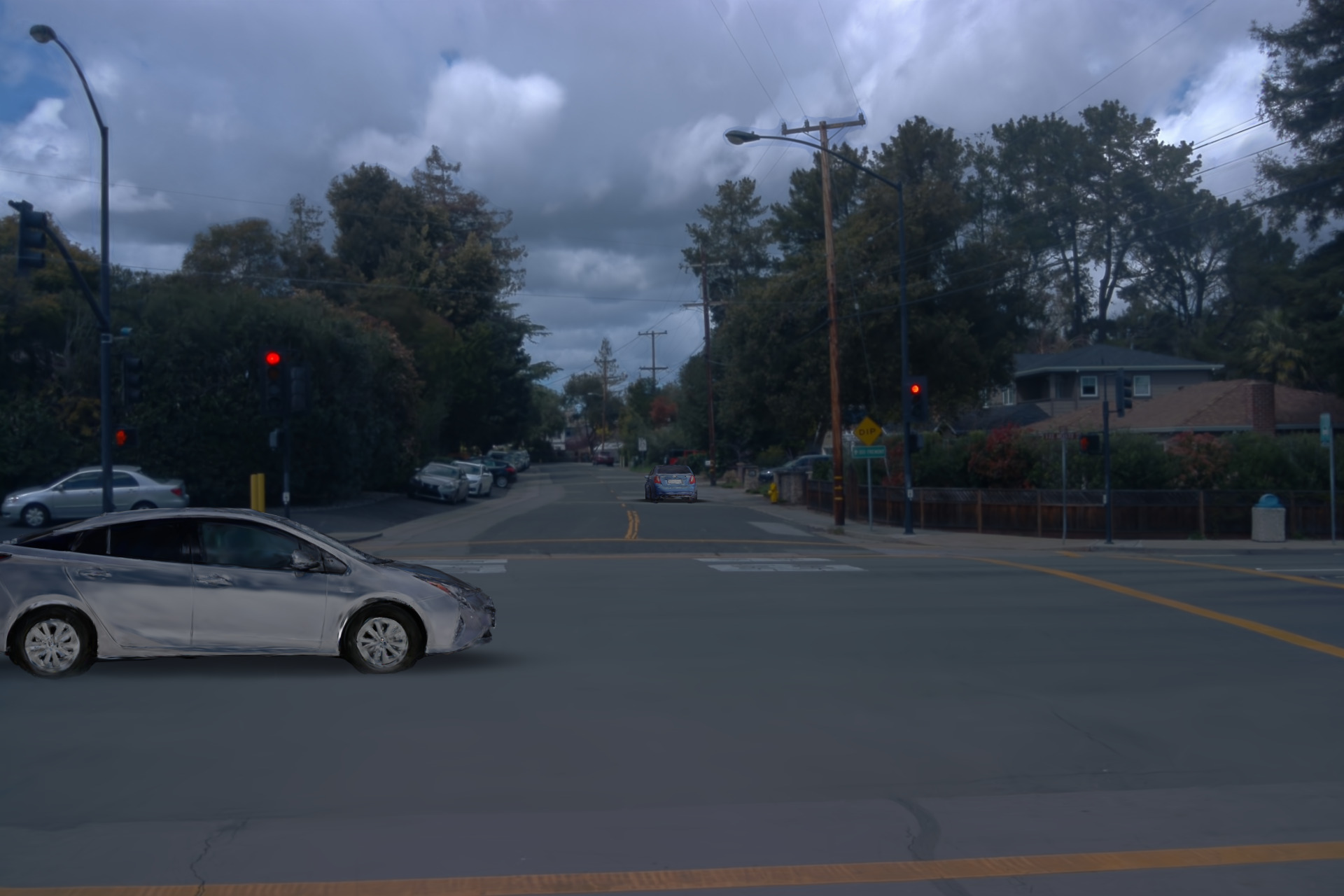} &
         \includegraphics[width=0.2\linewidth]{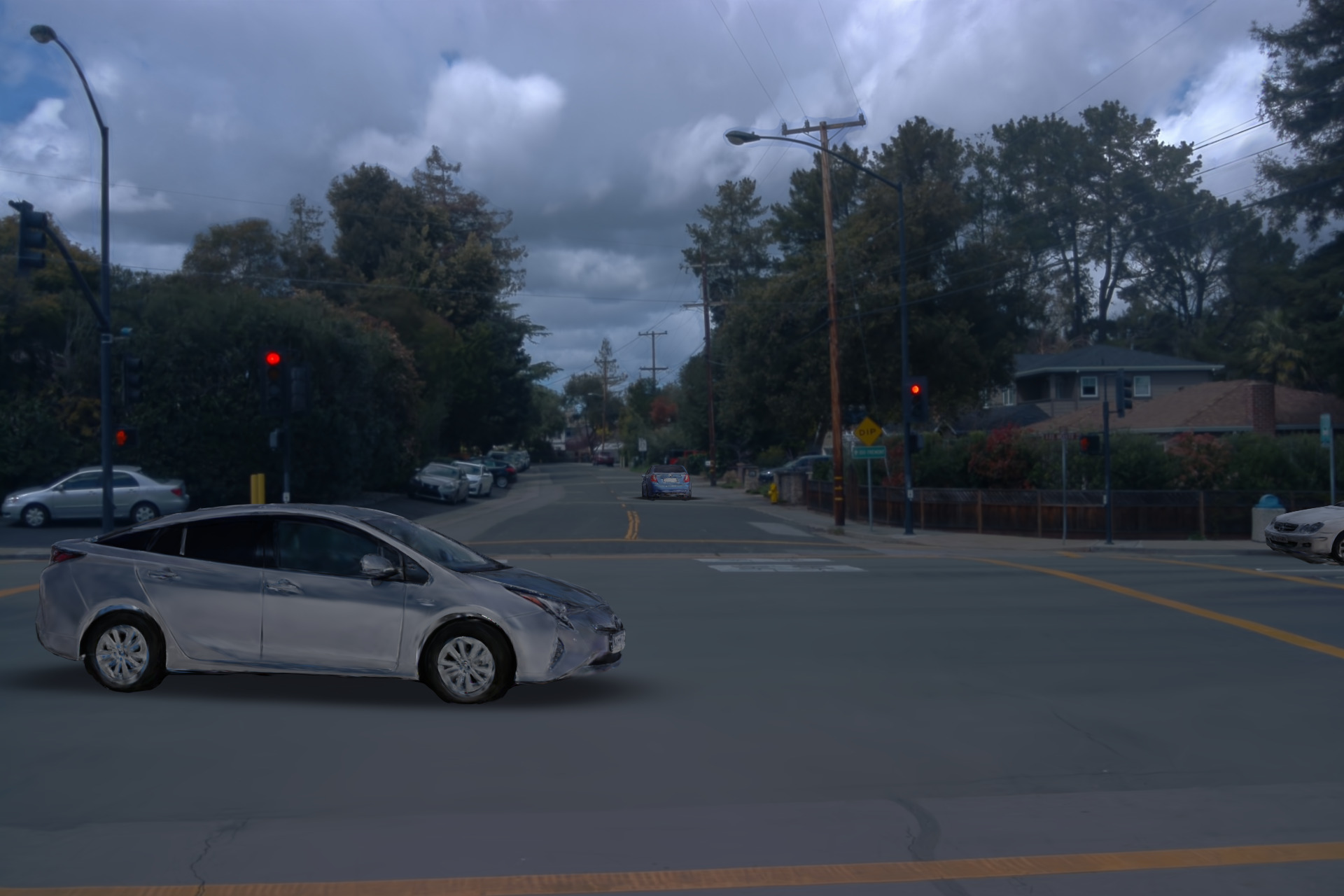} &
         \includegraphics[width=0.2\linewidth]{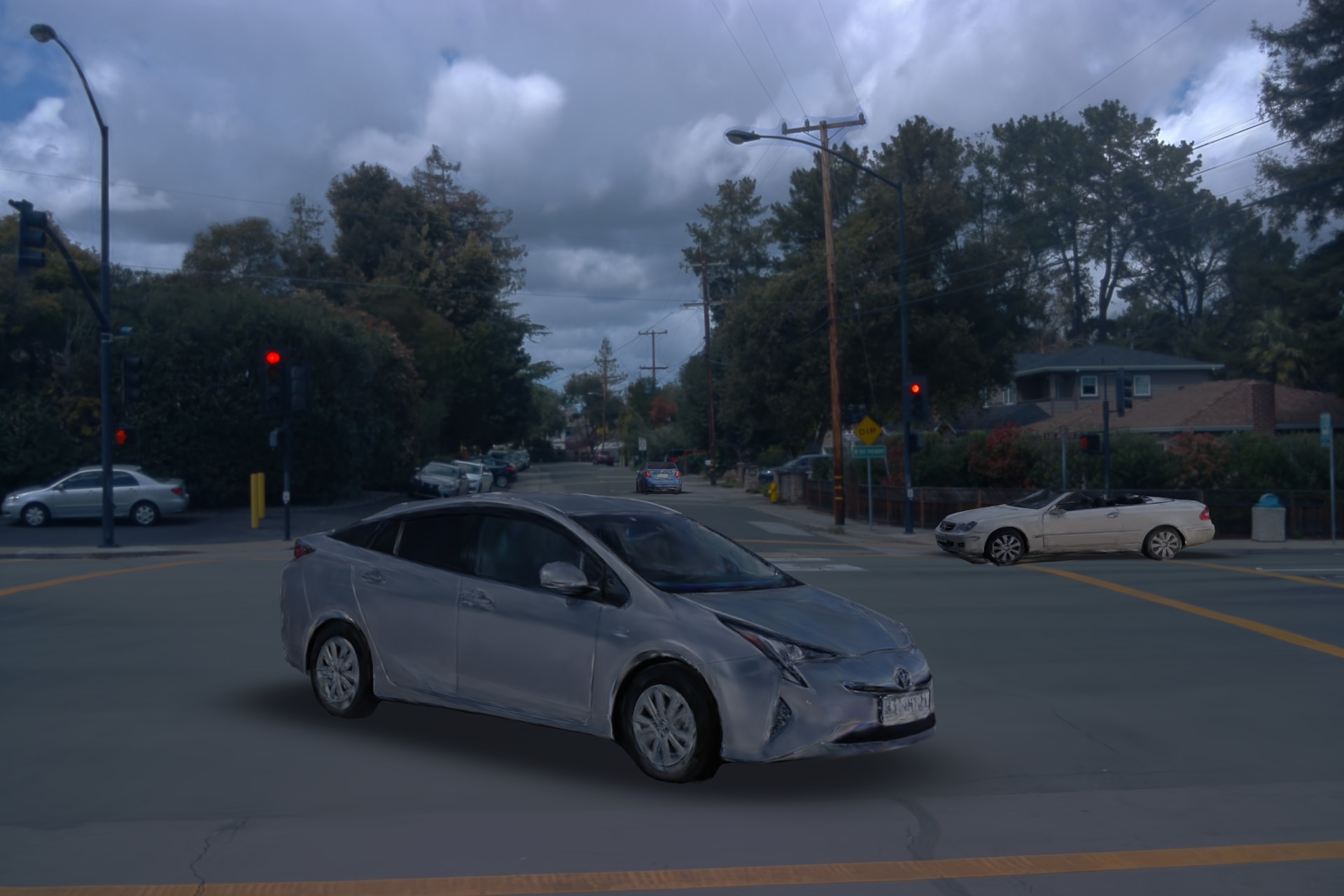} &
         \includegraphics[width=0.2\linewidth]{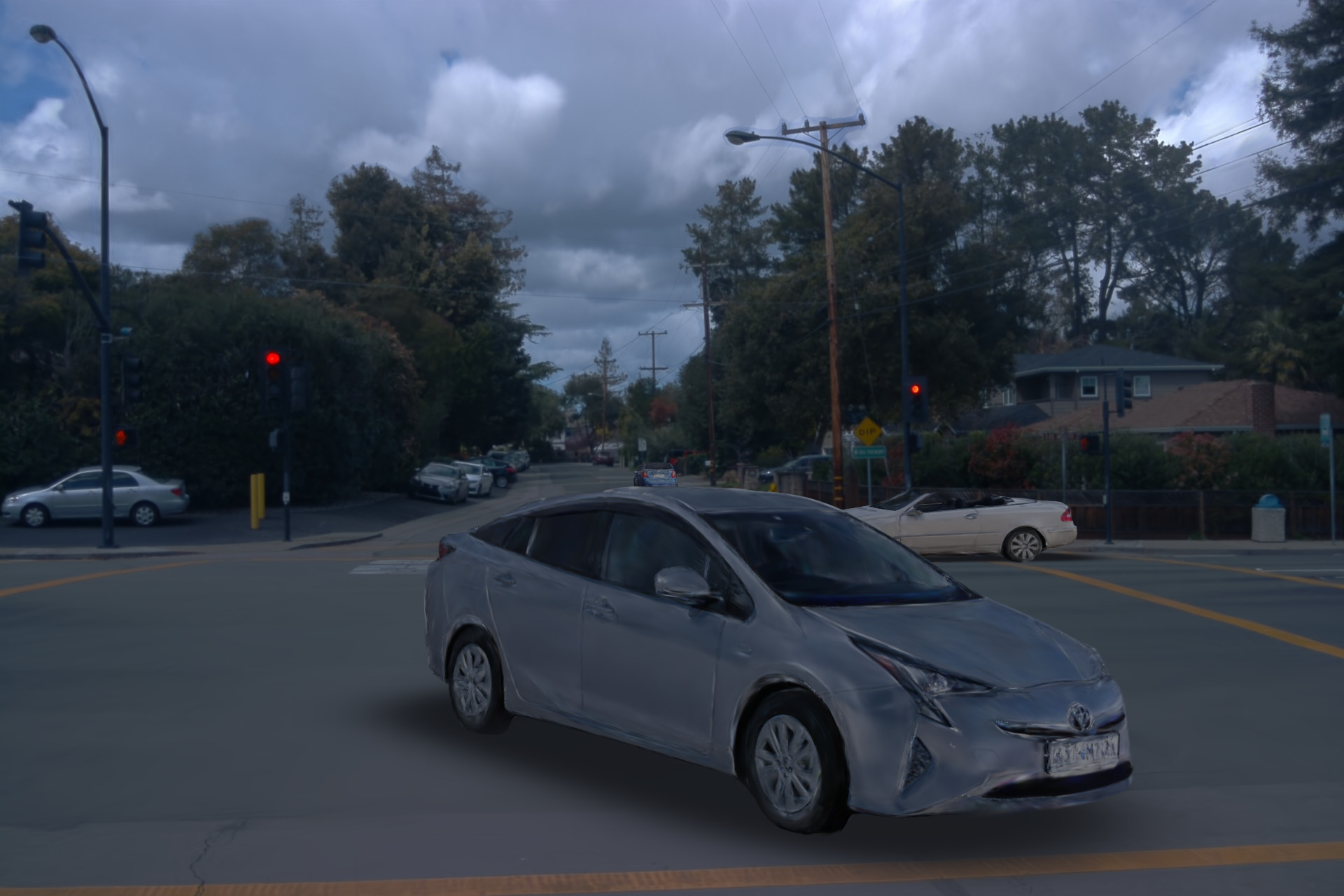} \\

         \rotatebox{90}{\parbox{2cm}{\centering\textbf{Original}}} &
         \includegraphics[width=0.2\linewidth]{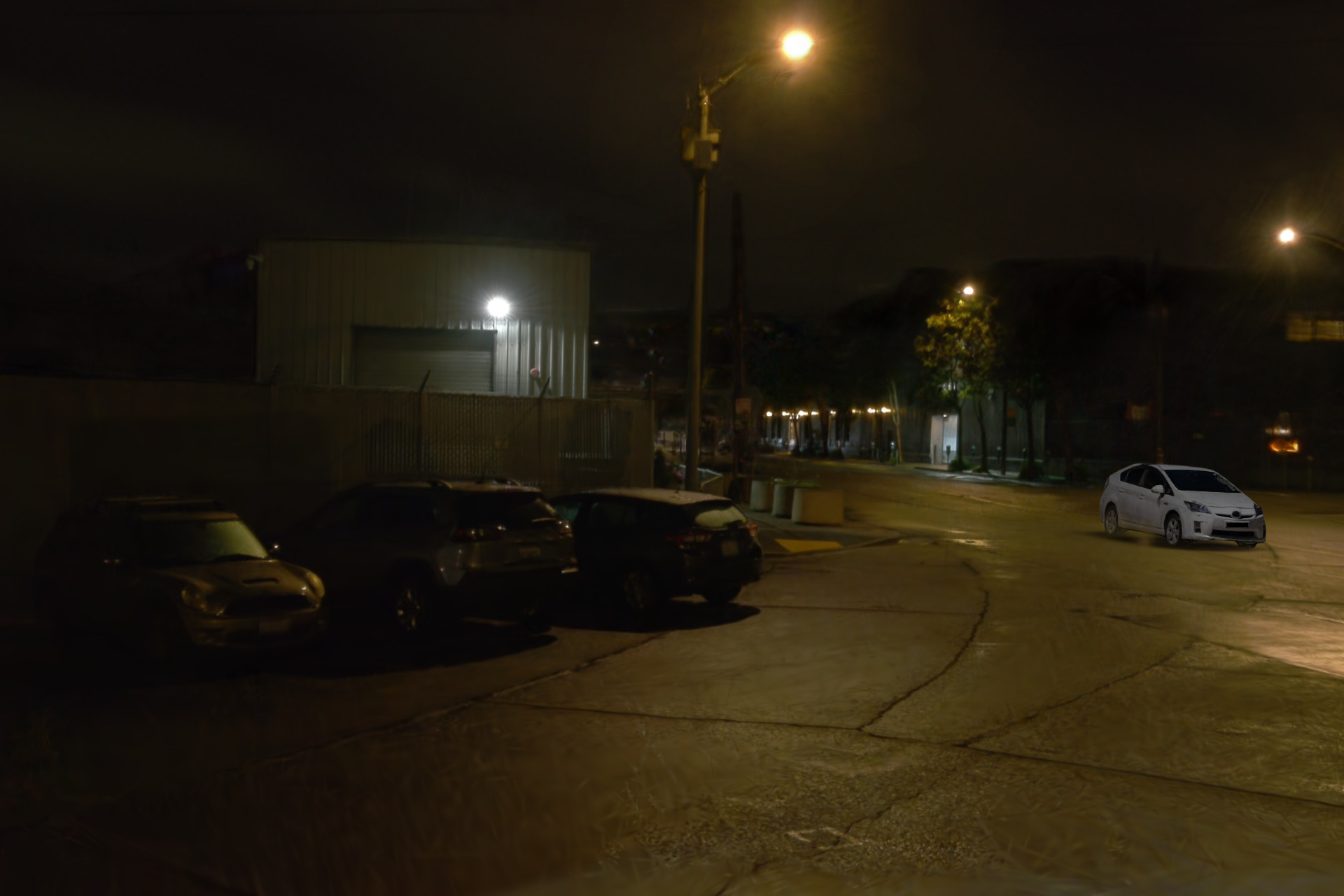} &
         \includegraphics[width=0.2\linewidth]{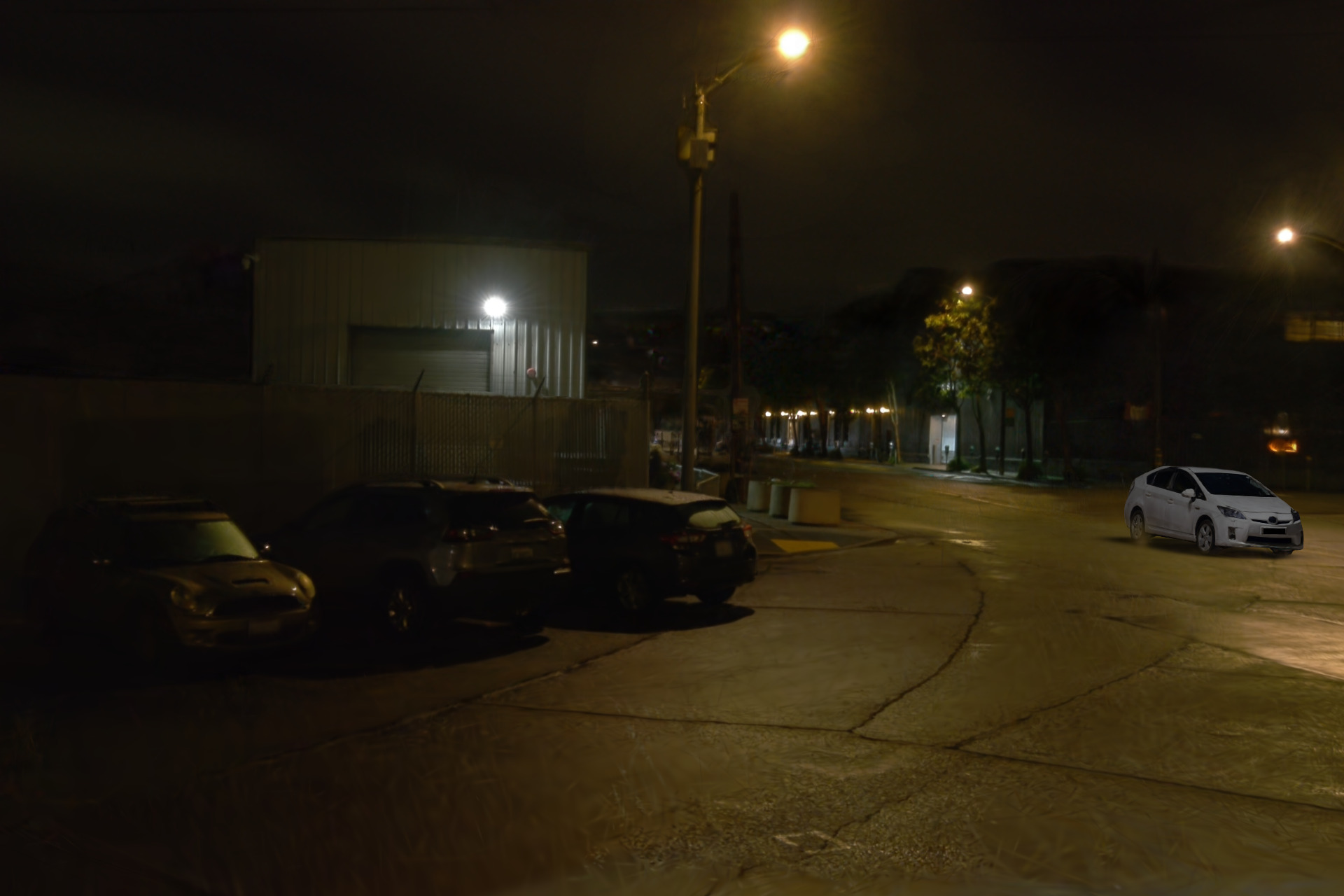} &
         \includegraphics[width=0.2\linewidth]{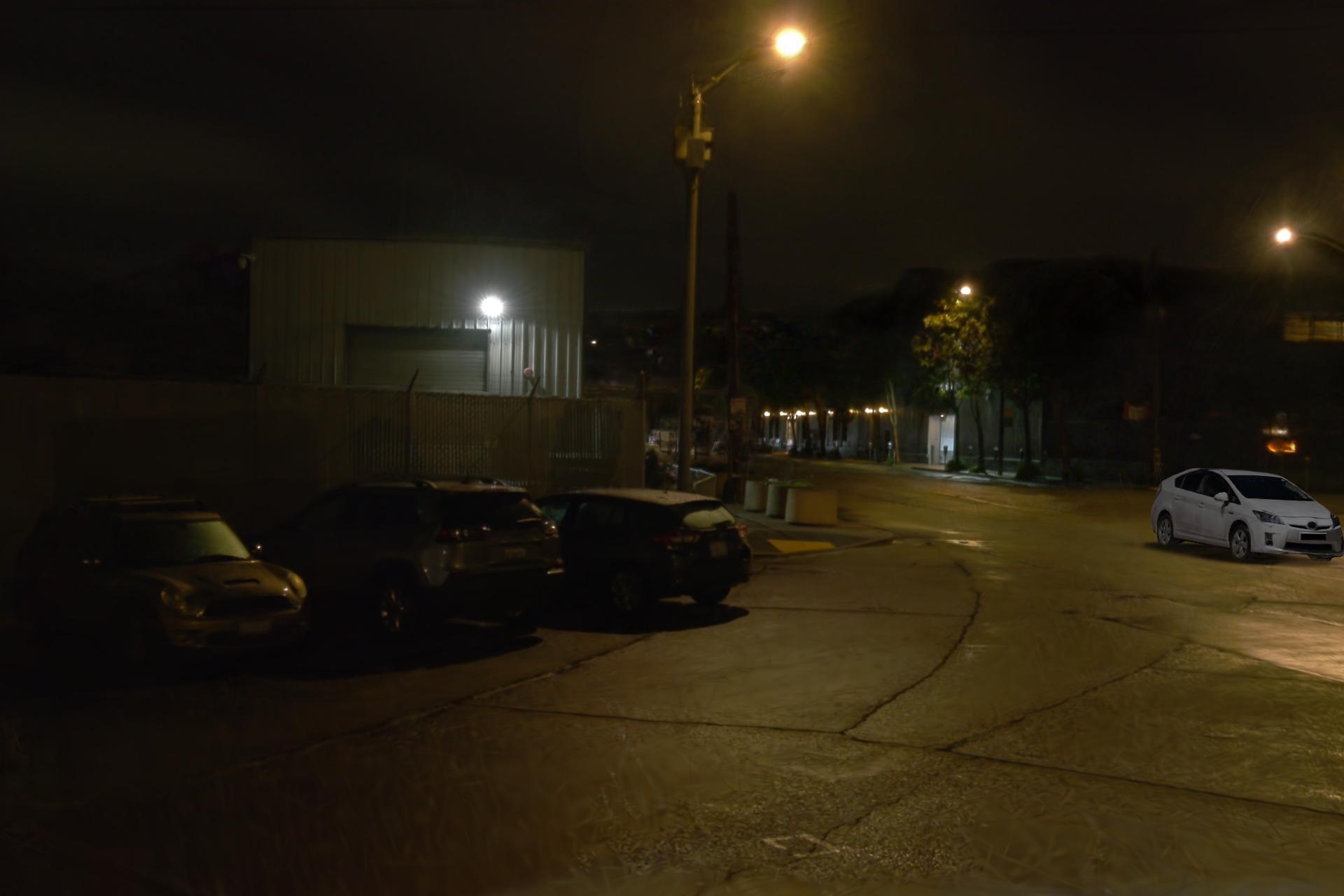} &
         \includegraphics[width=0.2\linewidth]{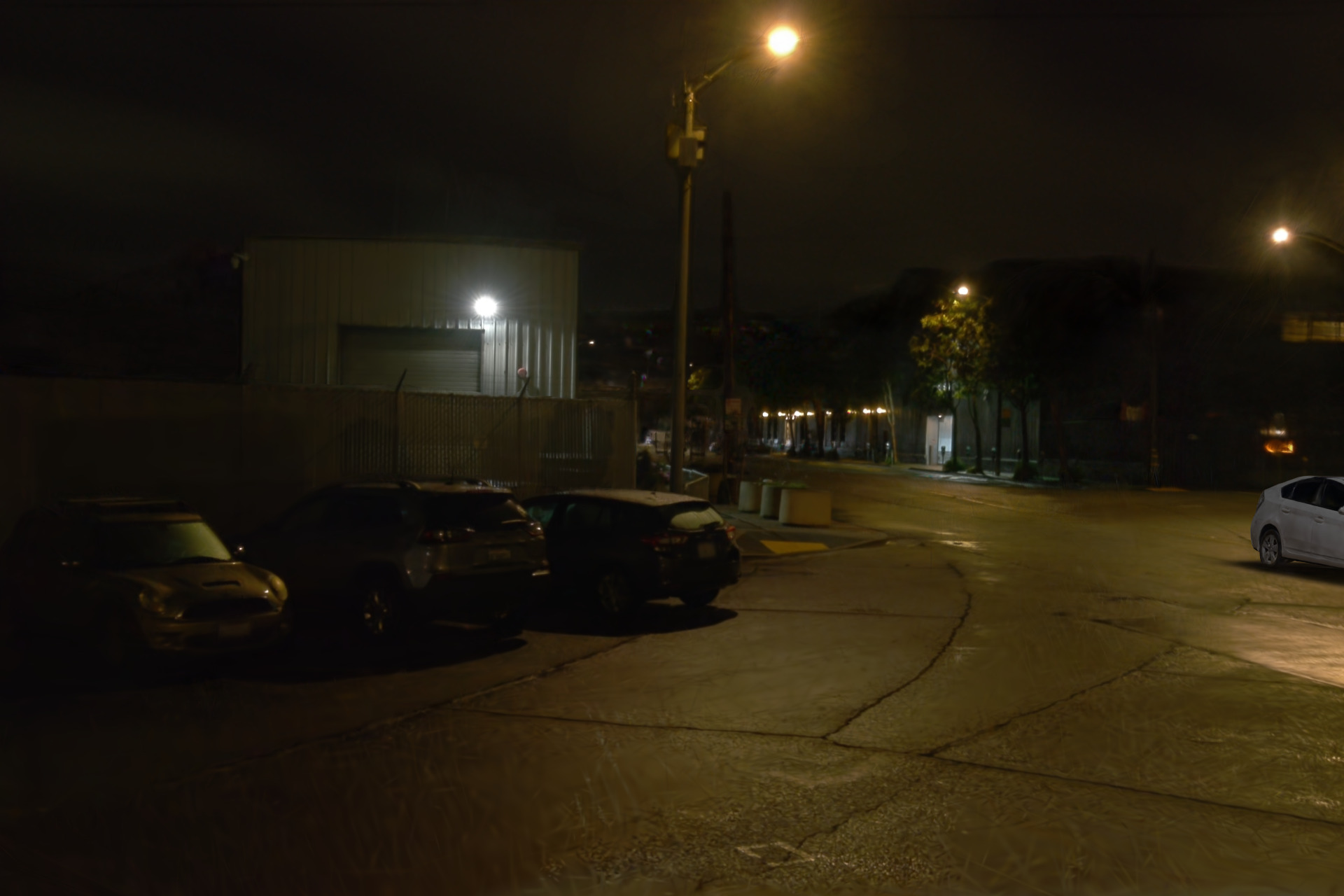} \\
         \rotatebox{90}{\parbox{2cm}{\centering\textbf{Modified}}} &
         \includegraphics[width=0.2\linewidth]{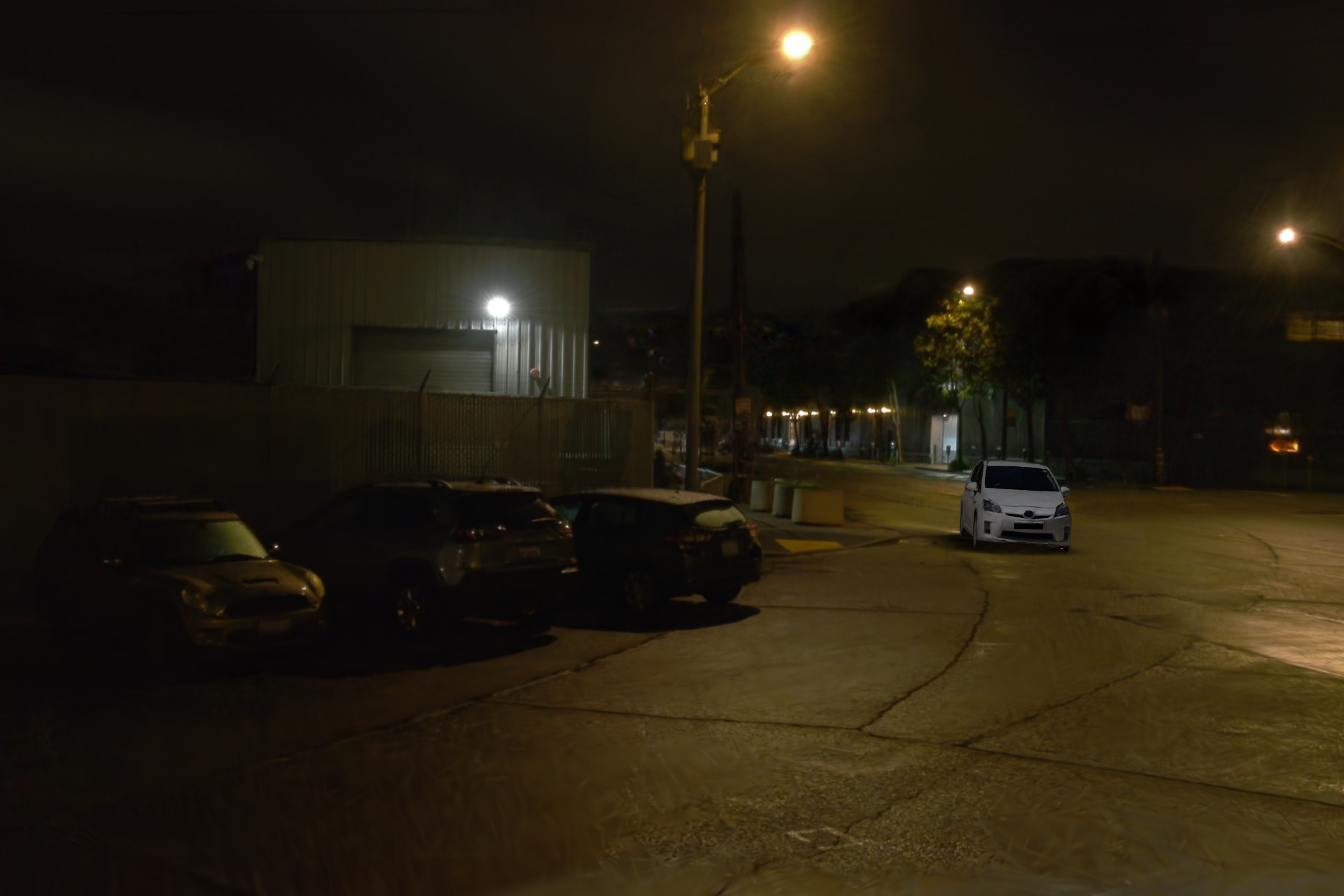} &
         \includegraphics[width=0.2\linewidth]{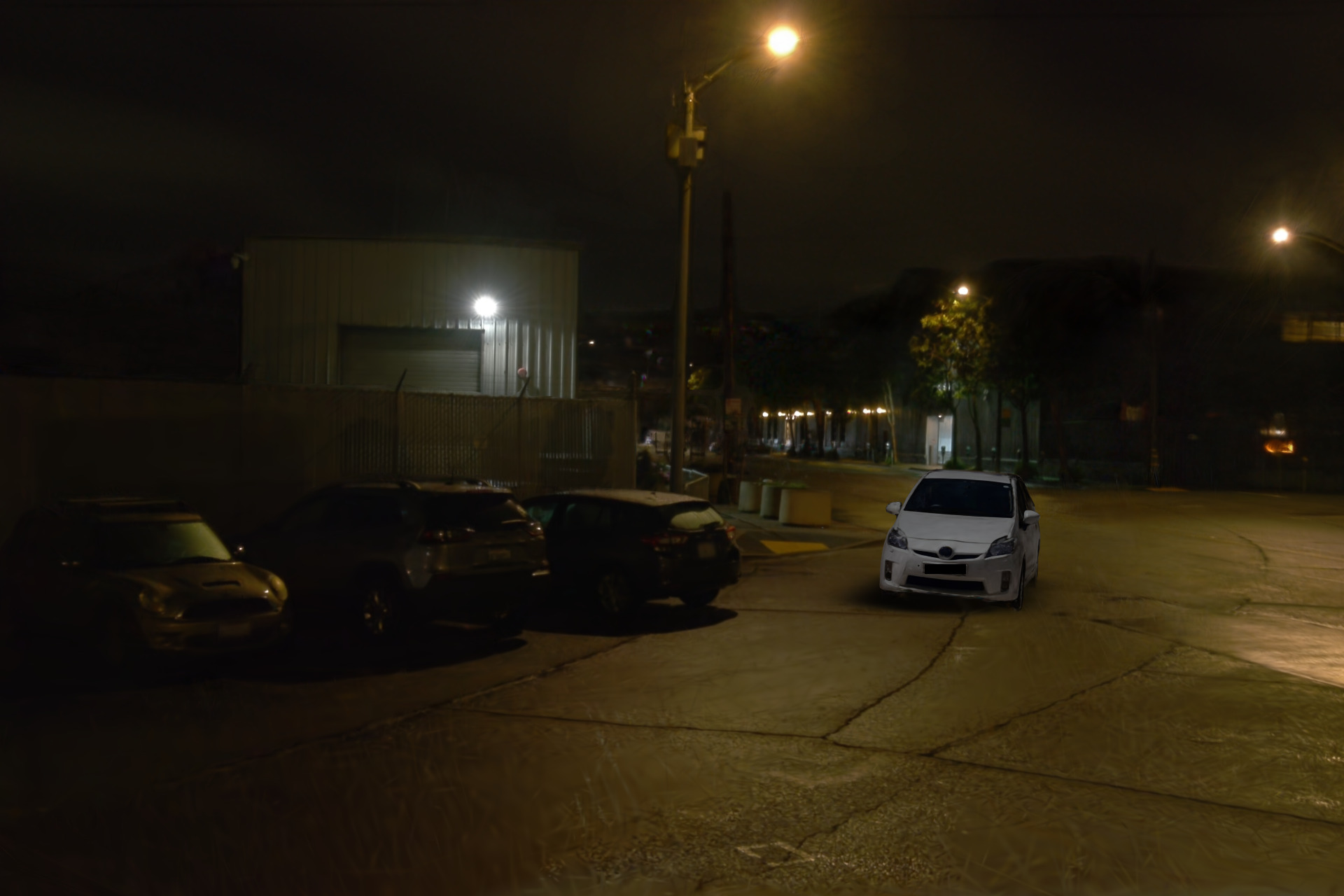} &
         \includegraphics[width=0.2\linewidth]{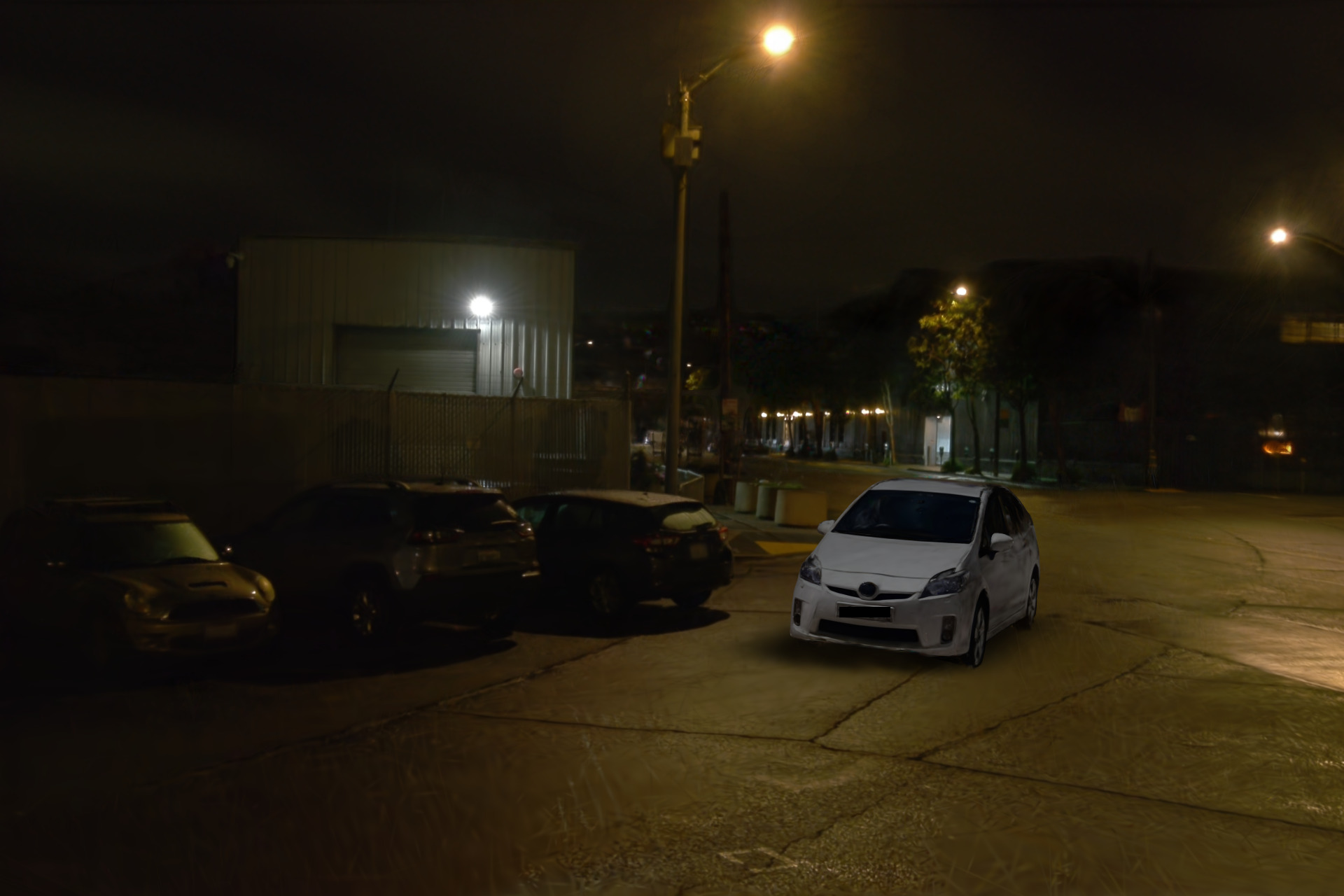} &
         \includegraphics[width=0.2\linewidth]{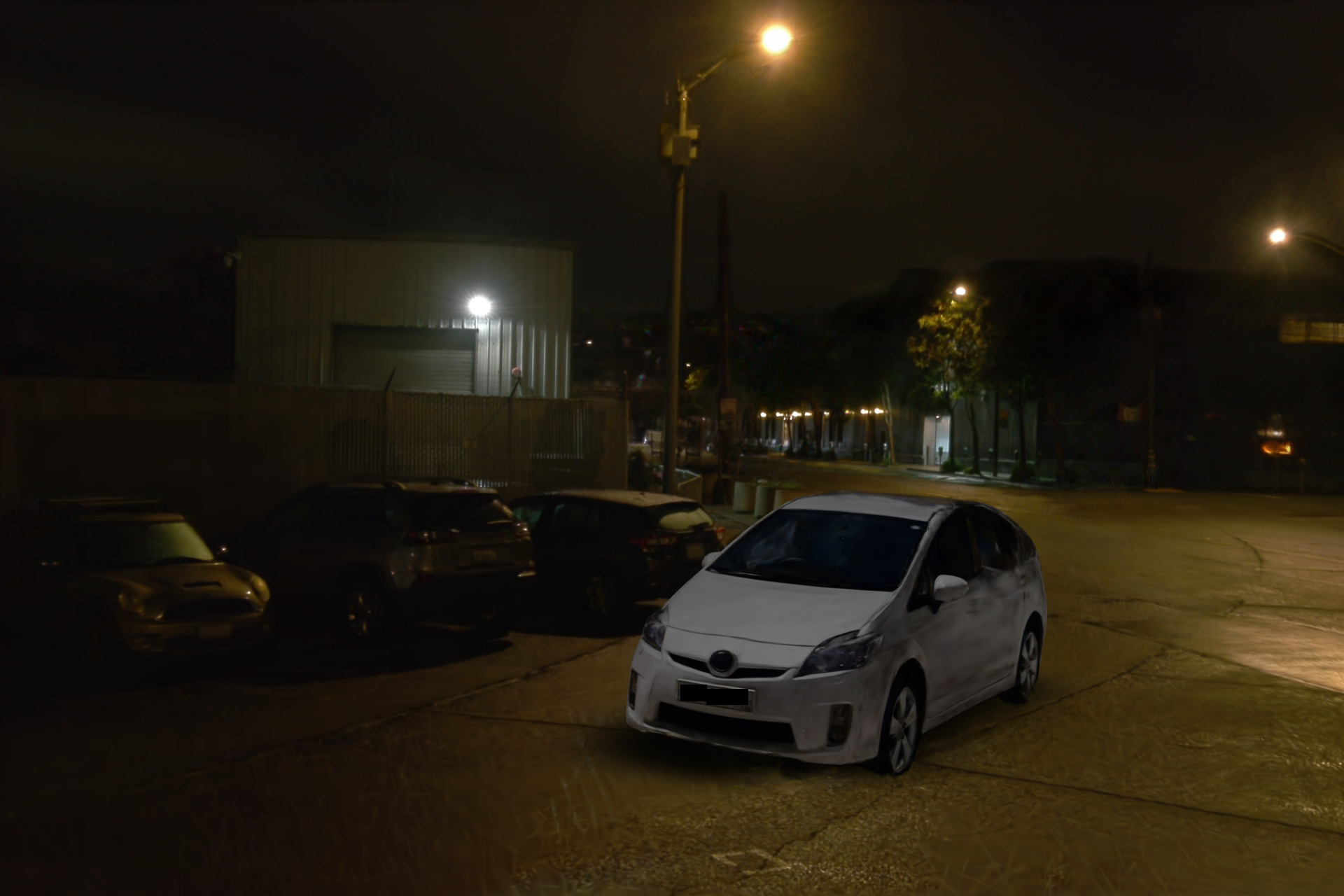} \\

         \rotatebox{90}{\parbox{2cm}{\centering\textbf{Original}}} &
         \includegraphics[width=0.2\linewidth]{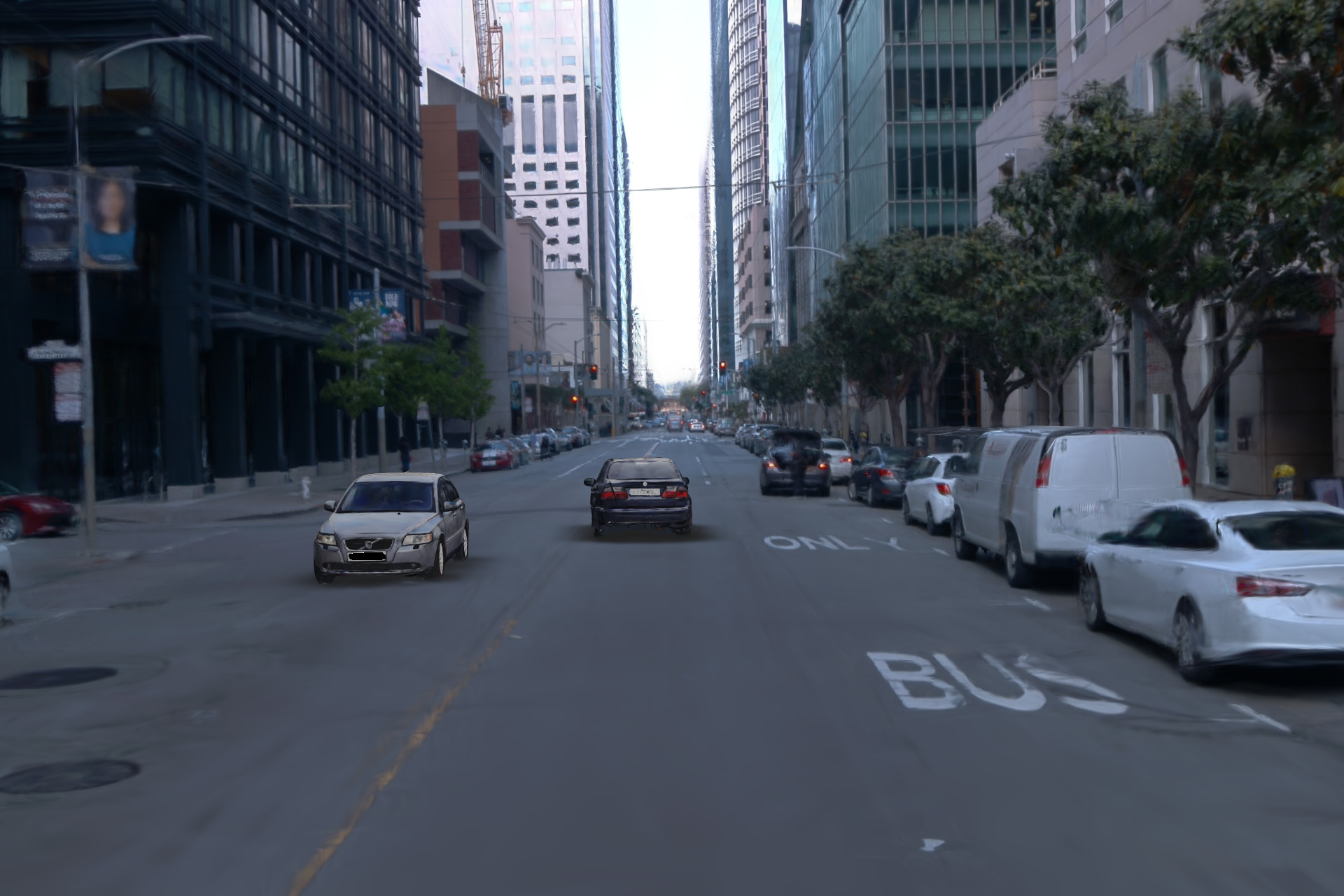} &
         \includegraphics[width=0.2\linewidth]{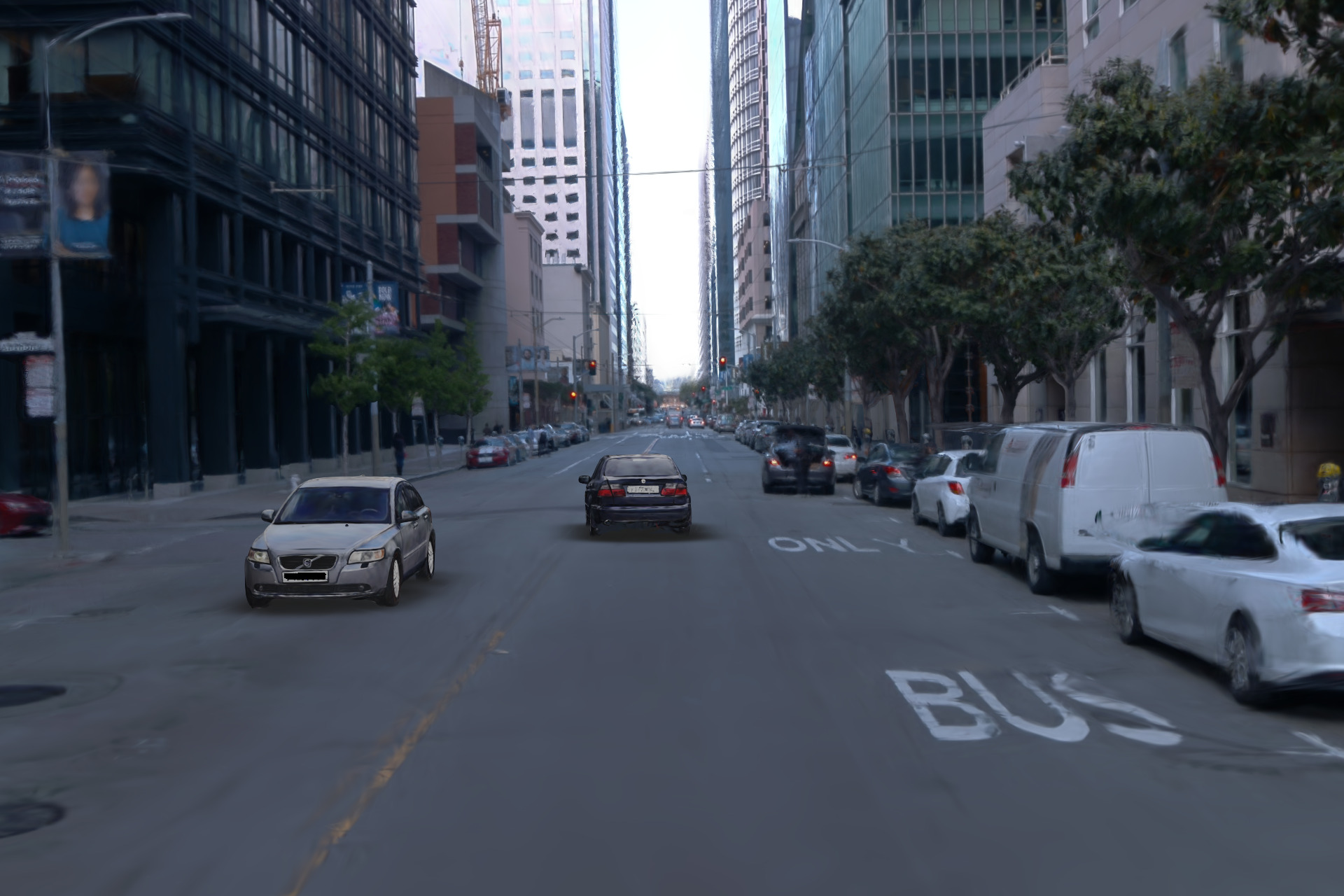} &
         \includegraphics[width=0.2\linewidth]{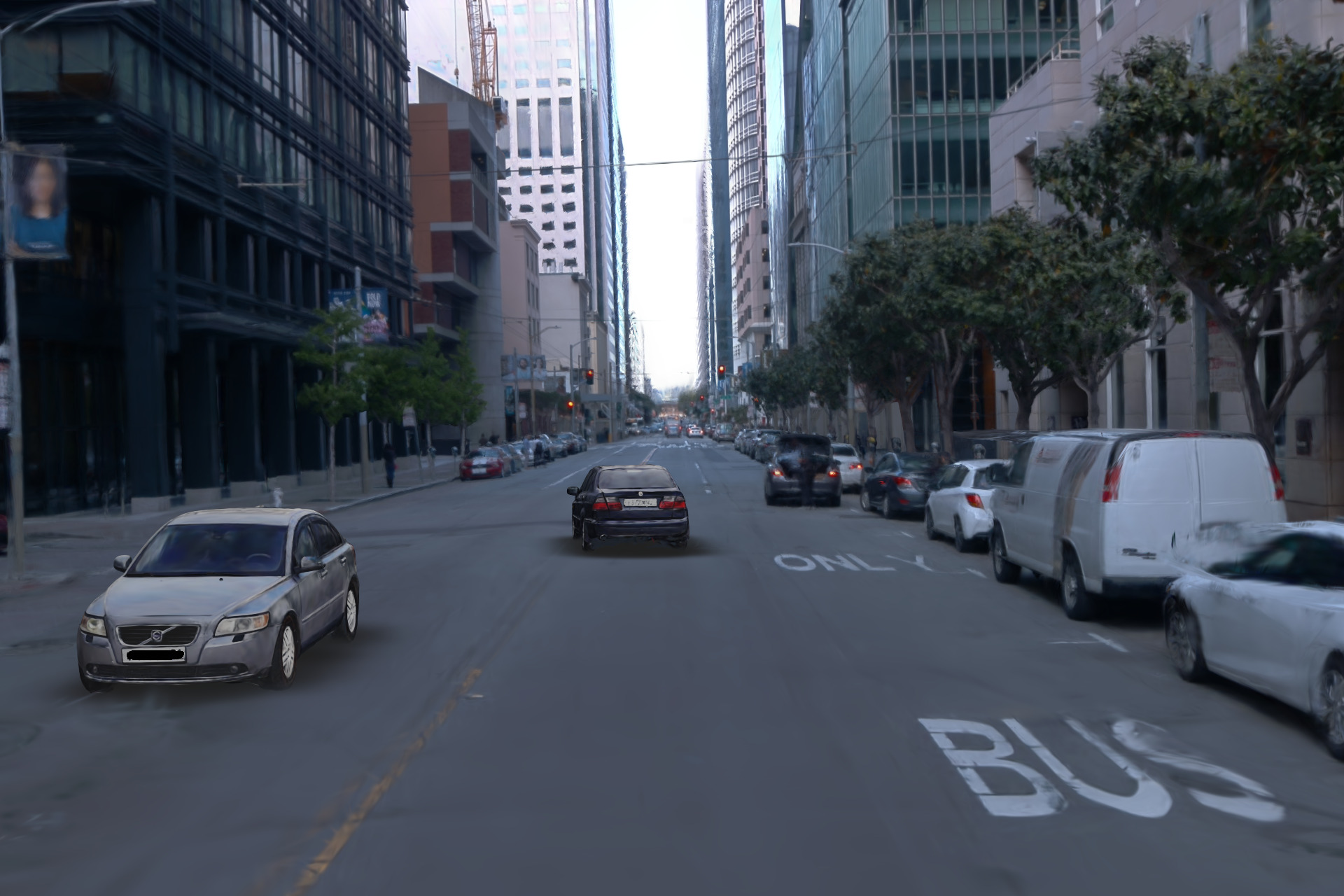} &
         \includegraphics[width=0.2\linewidth]{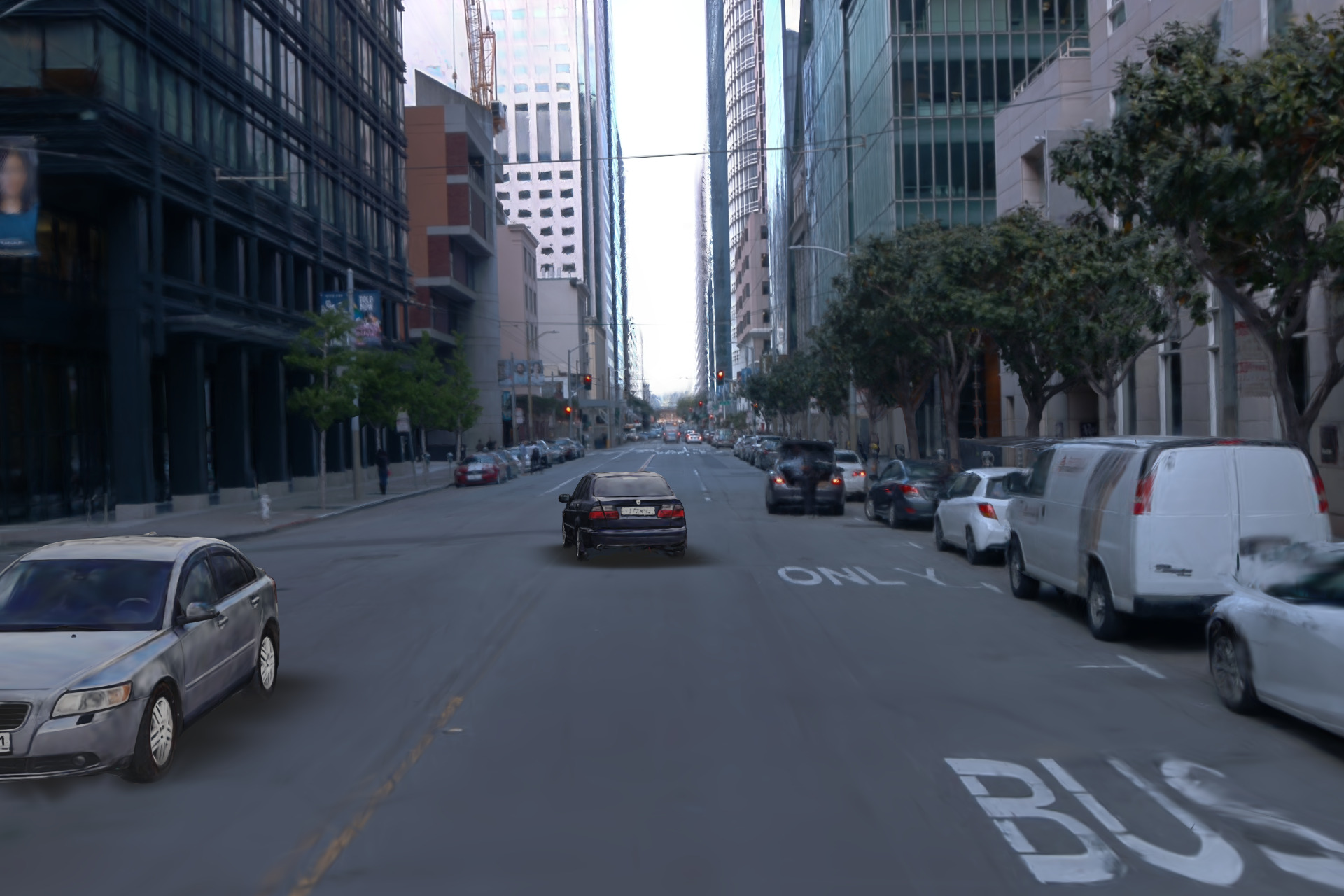} \\
         \rotatebox{90}{\parbox{2cm}{\centering\textbf{Modified}}} &
         \includegraphics[width=0.2\linewidth]{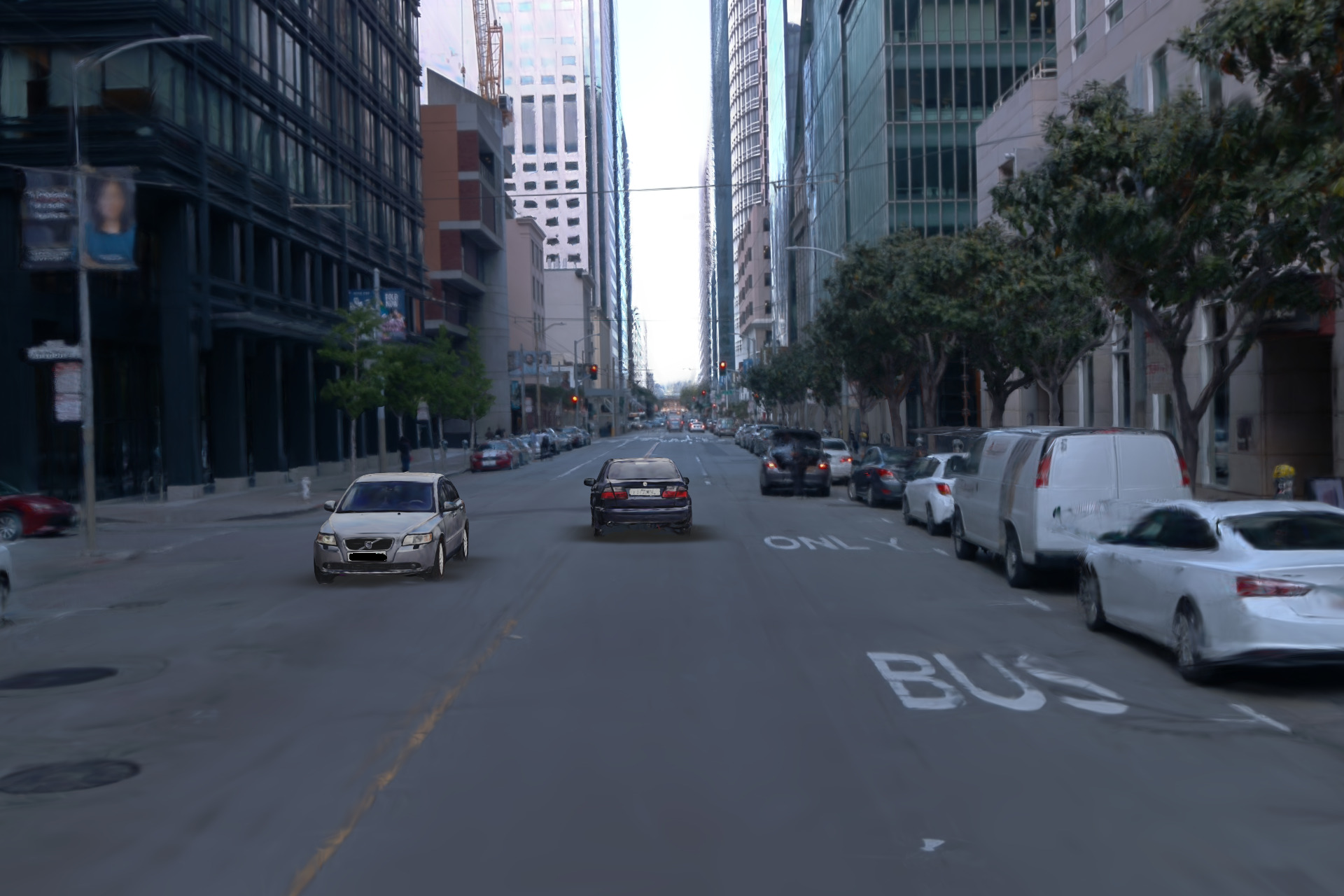} &
         \includegraphics[width=0.2\linewidth]{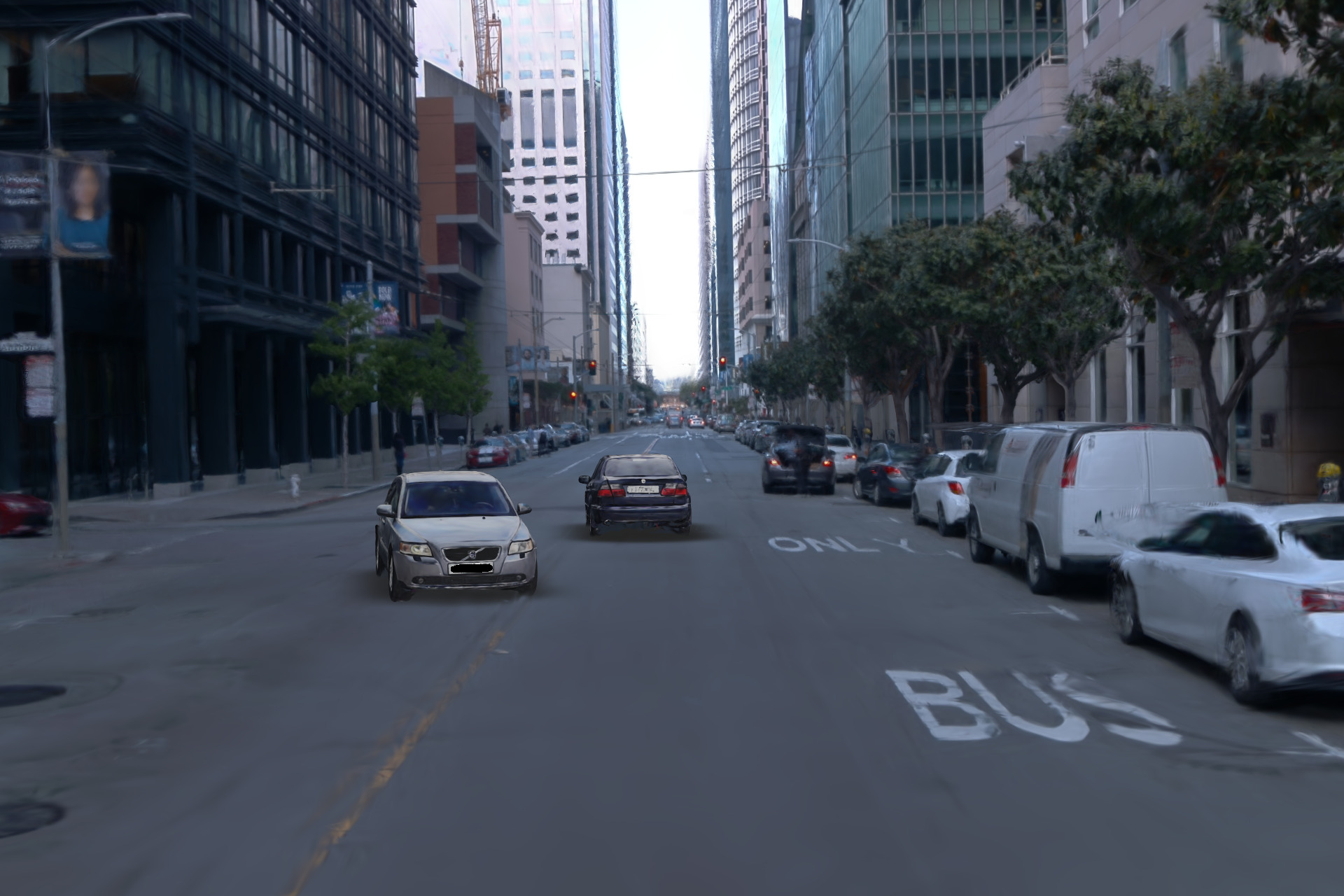} &
         \includegraphics[width=0.2\linewidth]{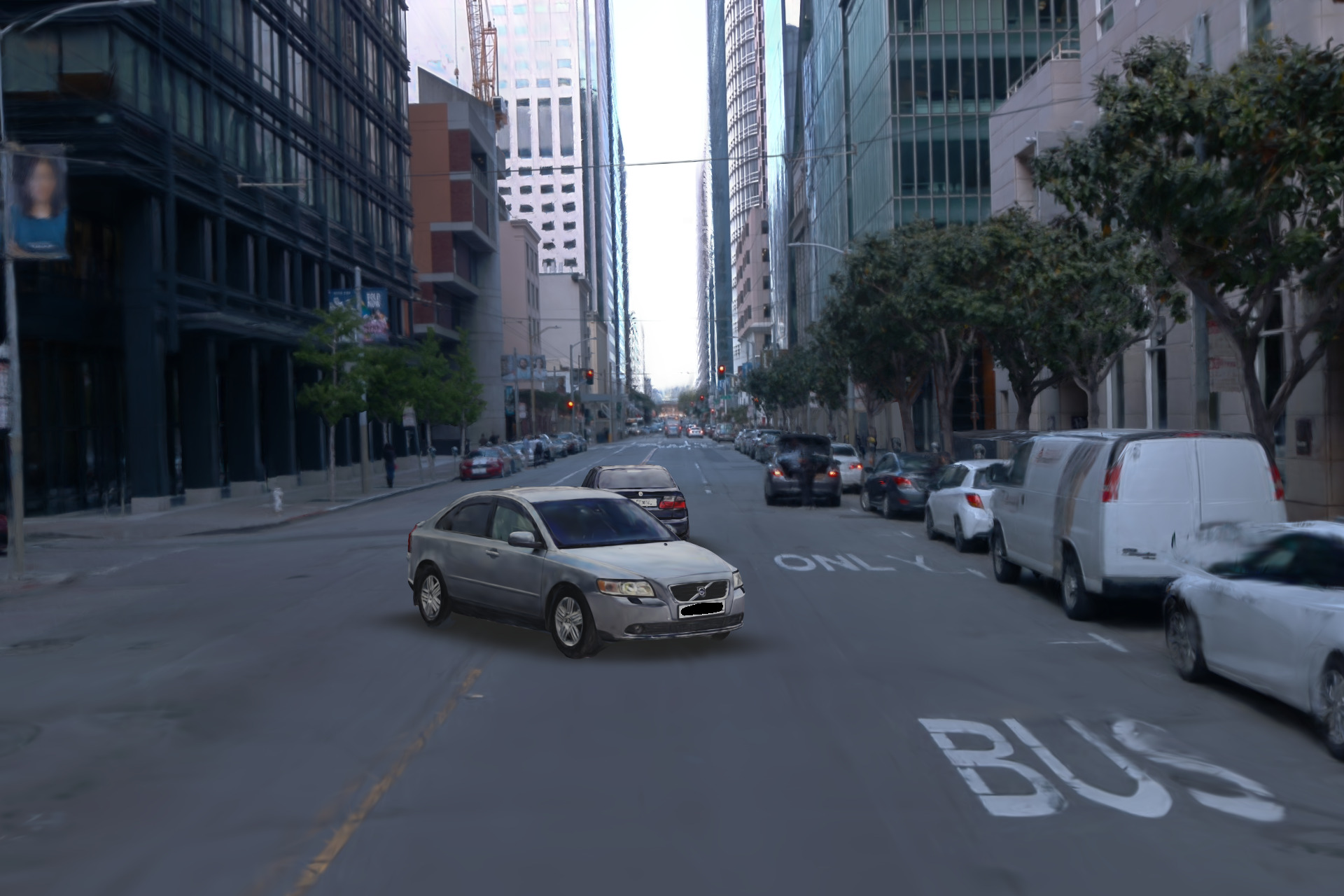} &
         \includegraphics[width=0.2\linewidth]{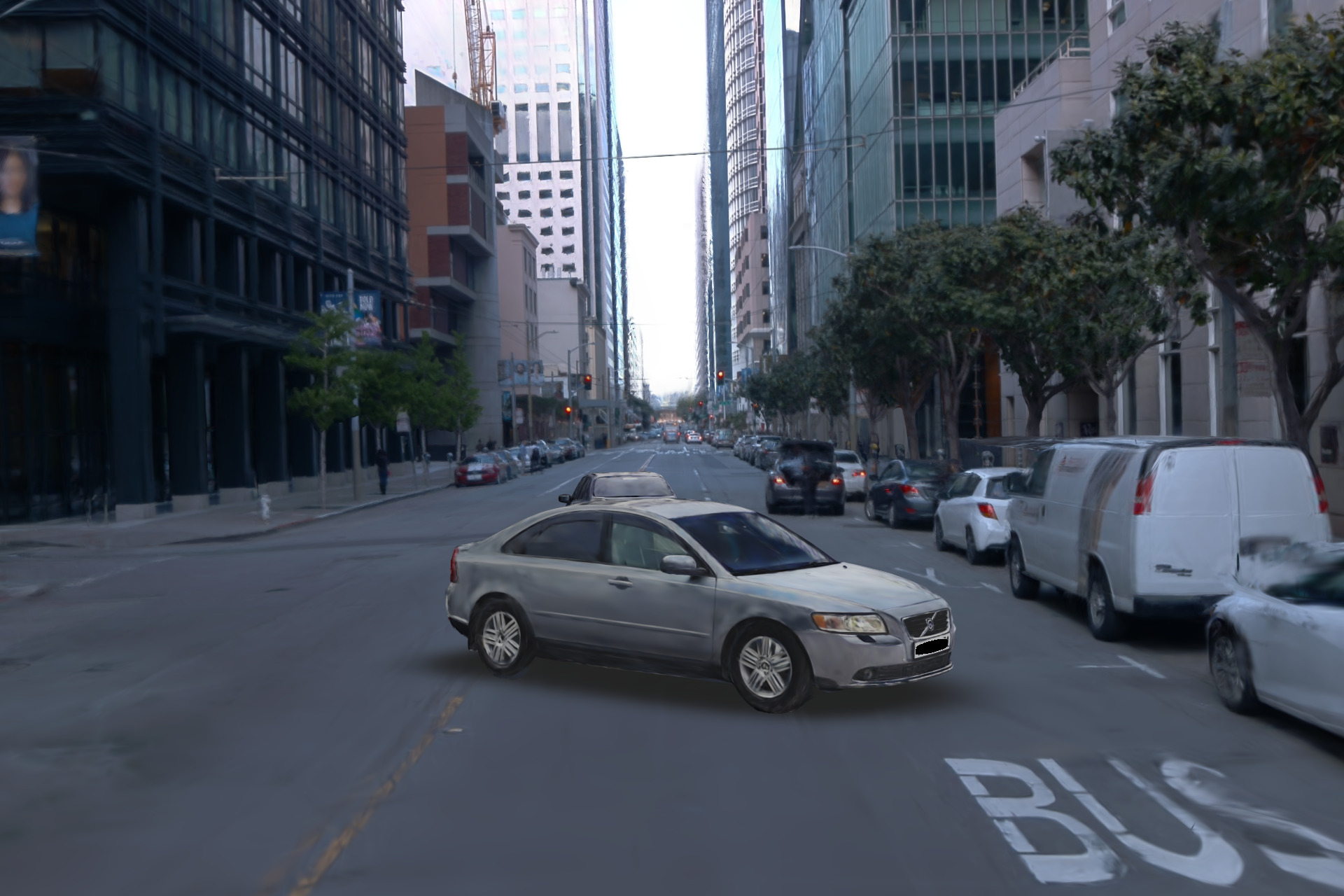} \\

         \rotatebox{90}{\parbox{2cm}{\centering\textbf{Original}}} &
         \includegraphics[width=0.2\linewidth]{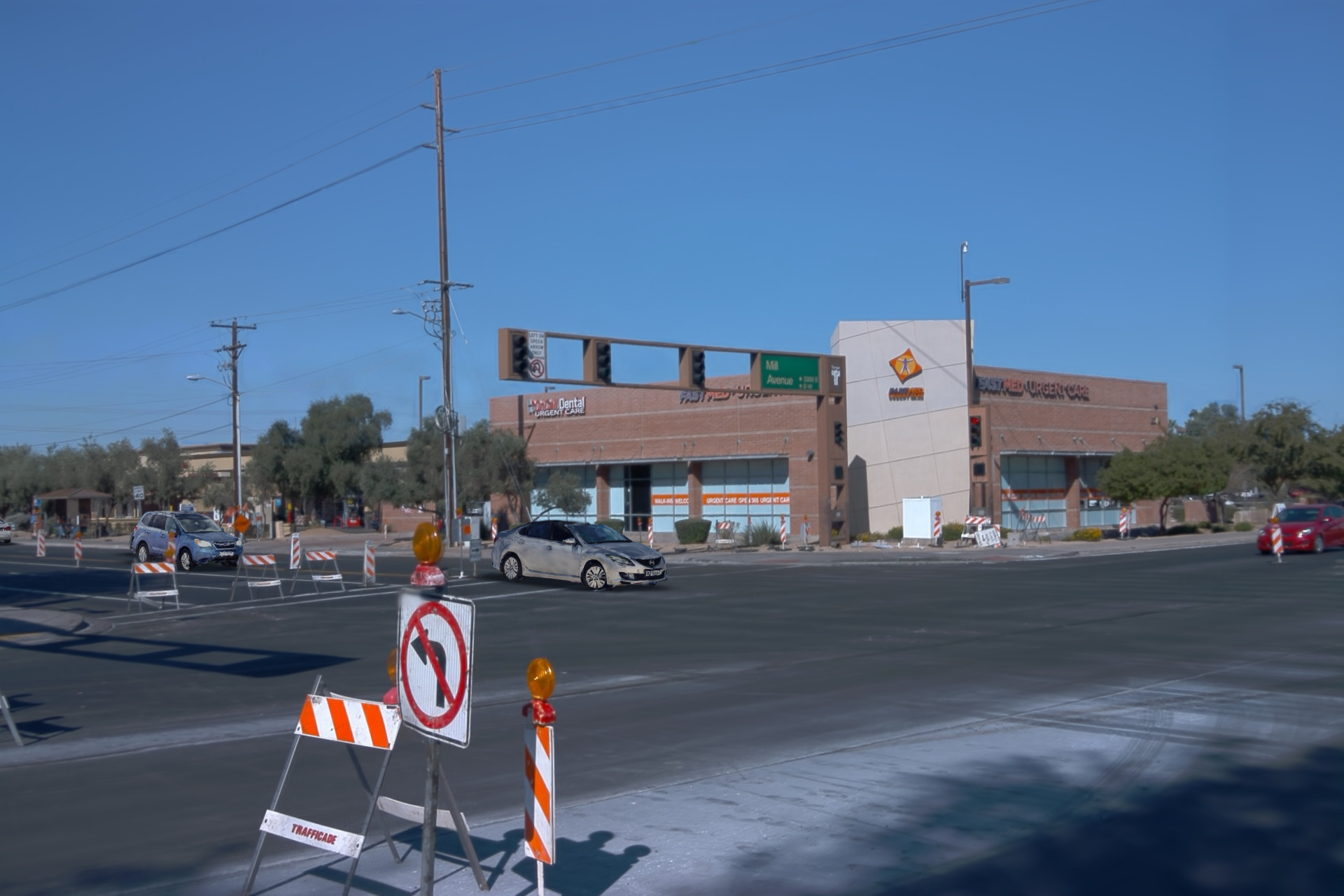} &
         \includegraphics[width=0.2\linewidth]{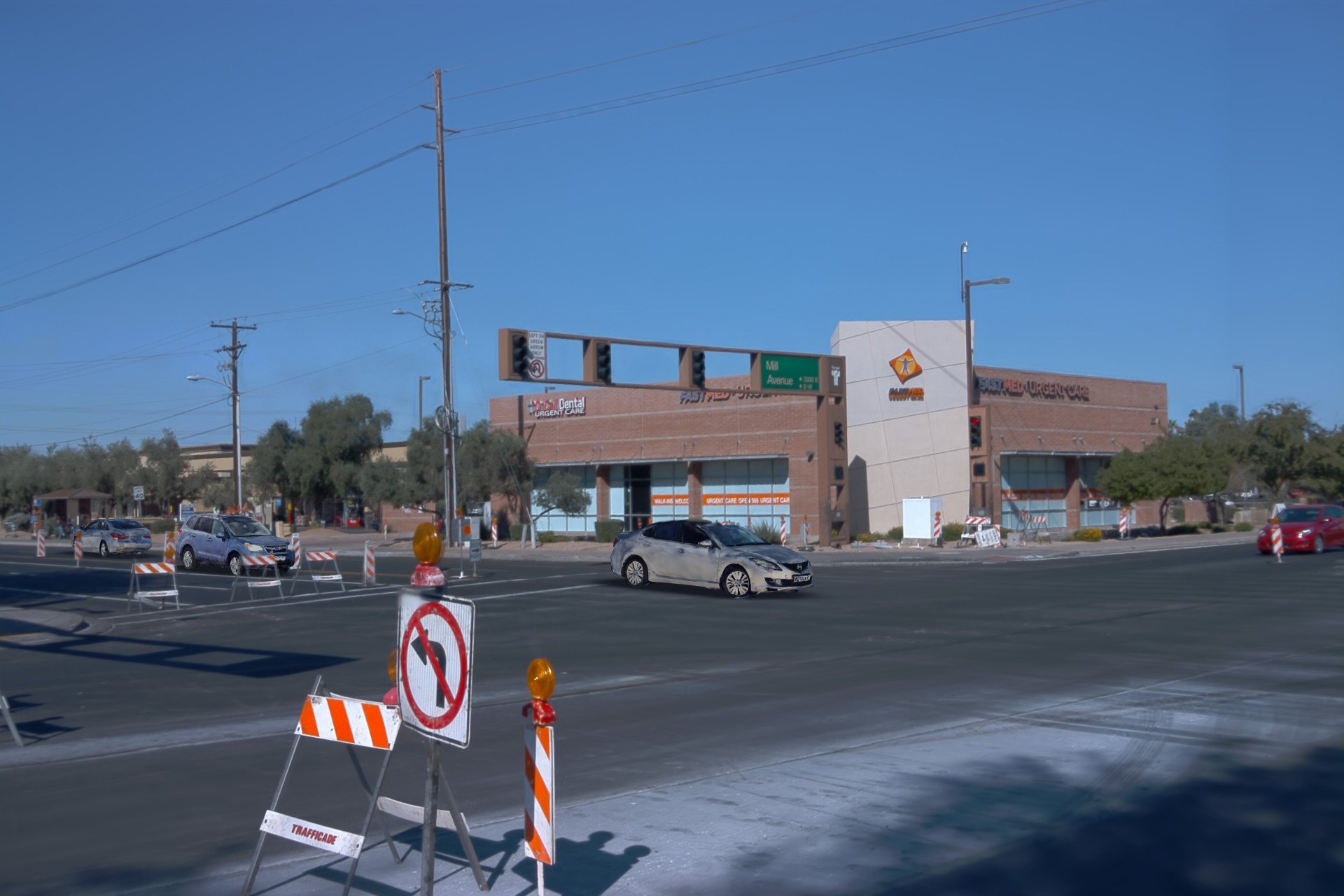} &
         \includegraphics[width=0.2\linewidth]{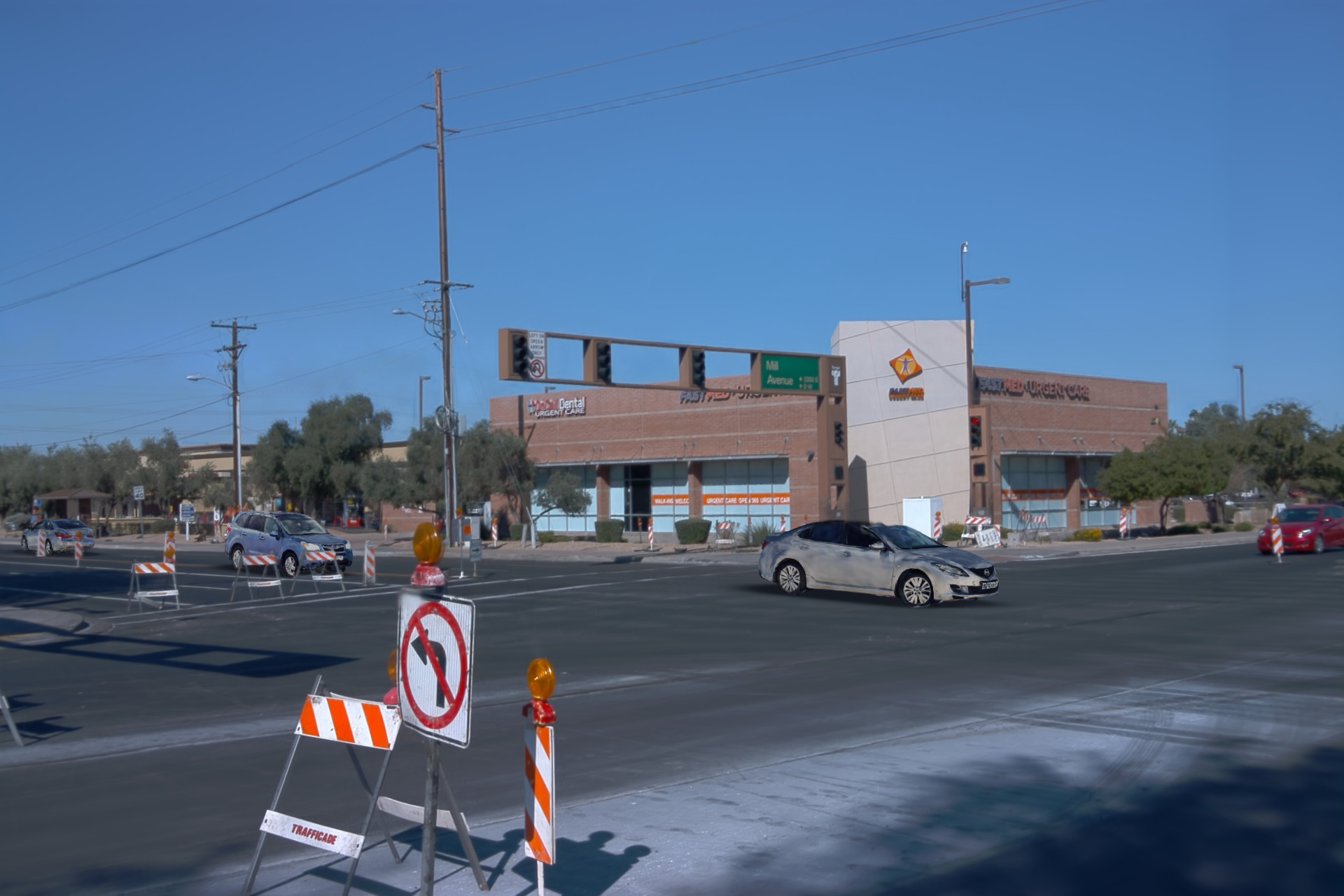} &
         \includegraphics[width=0.2\linewidth]{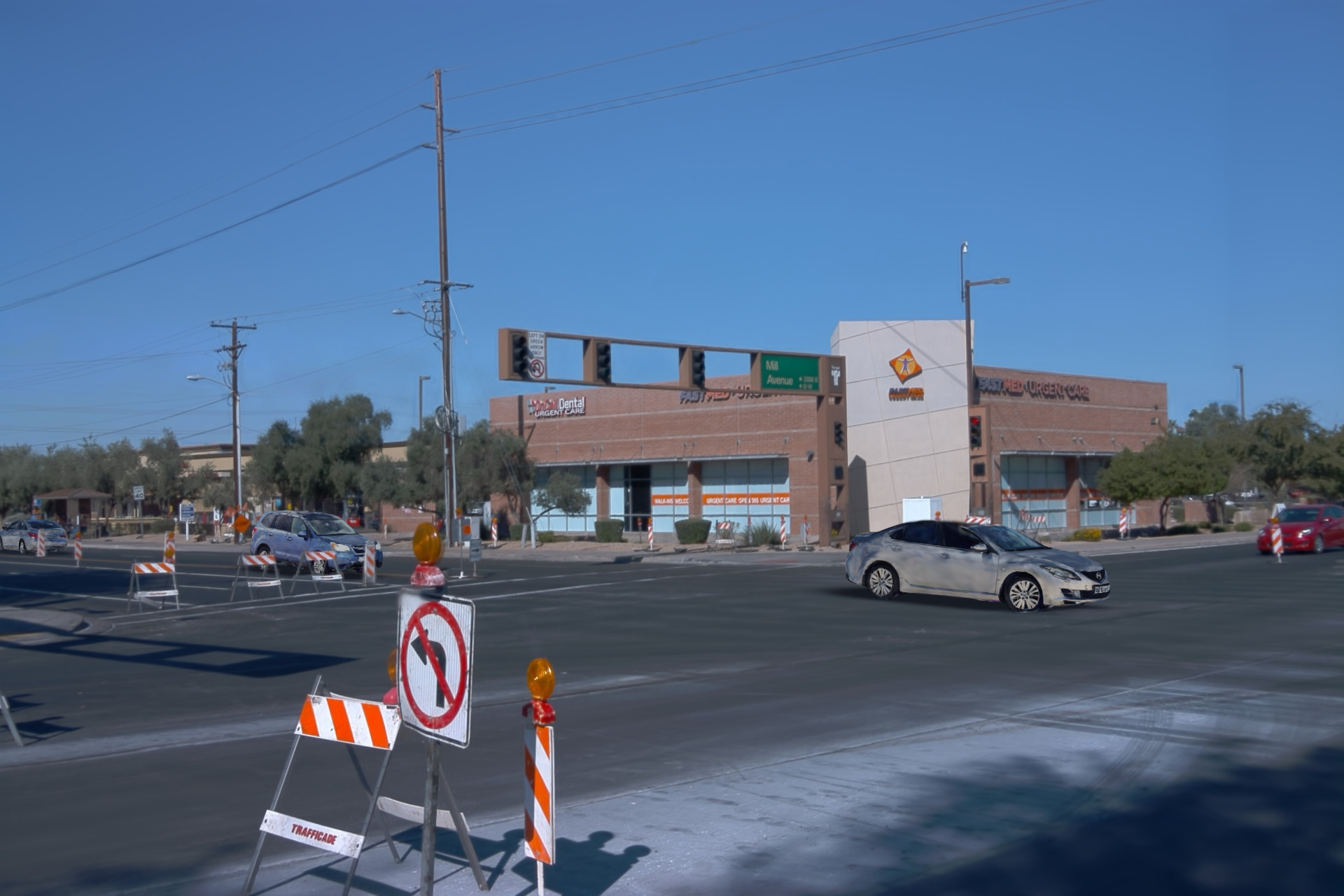} \\
         \rotatebox{90}{\parbox{2cm}{\centering\textbf{Modified}}} &
         \includegraphics[width=0.2\linewidth]{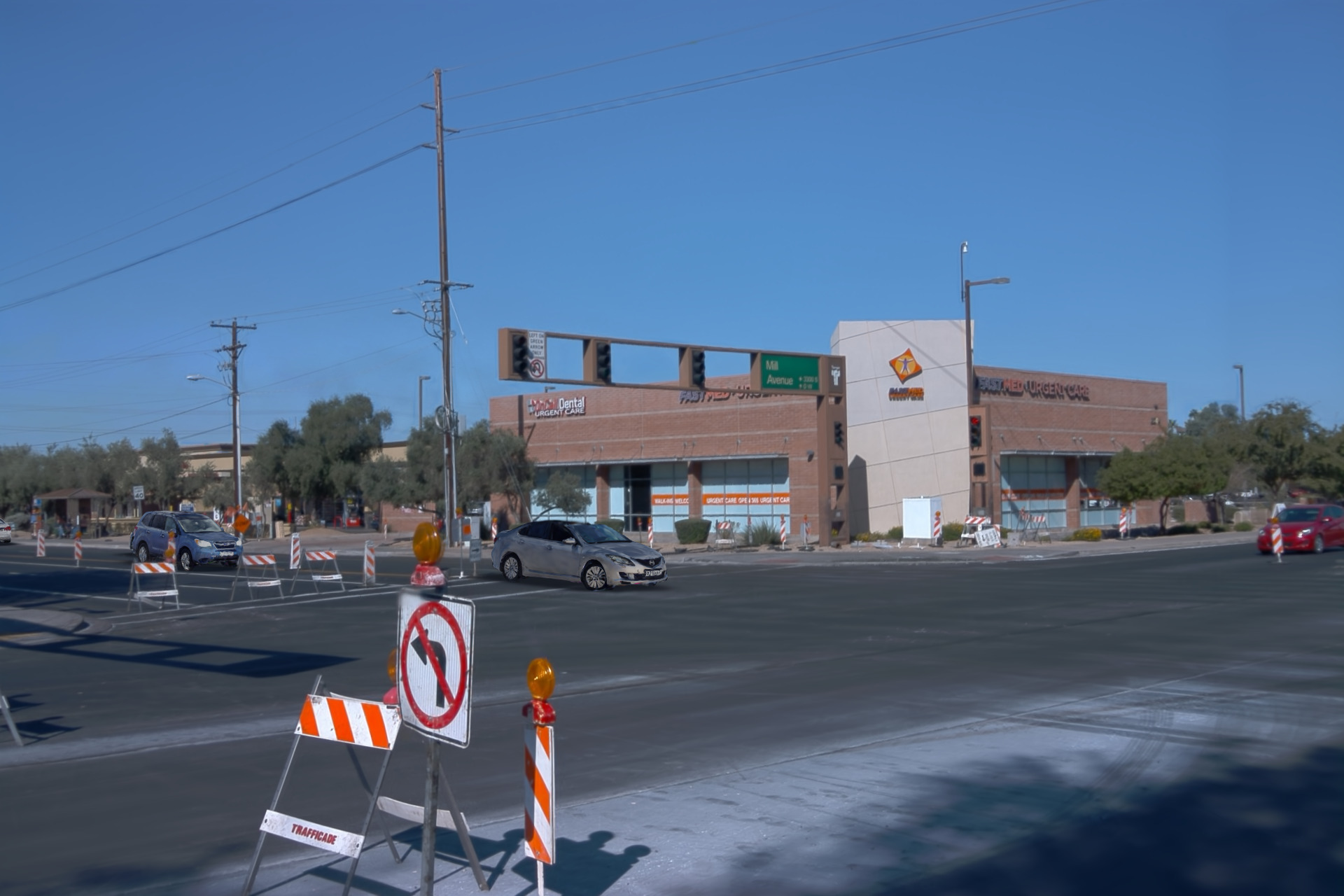} &
         \includegraphics[width=0.2\linewidth]{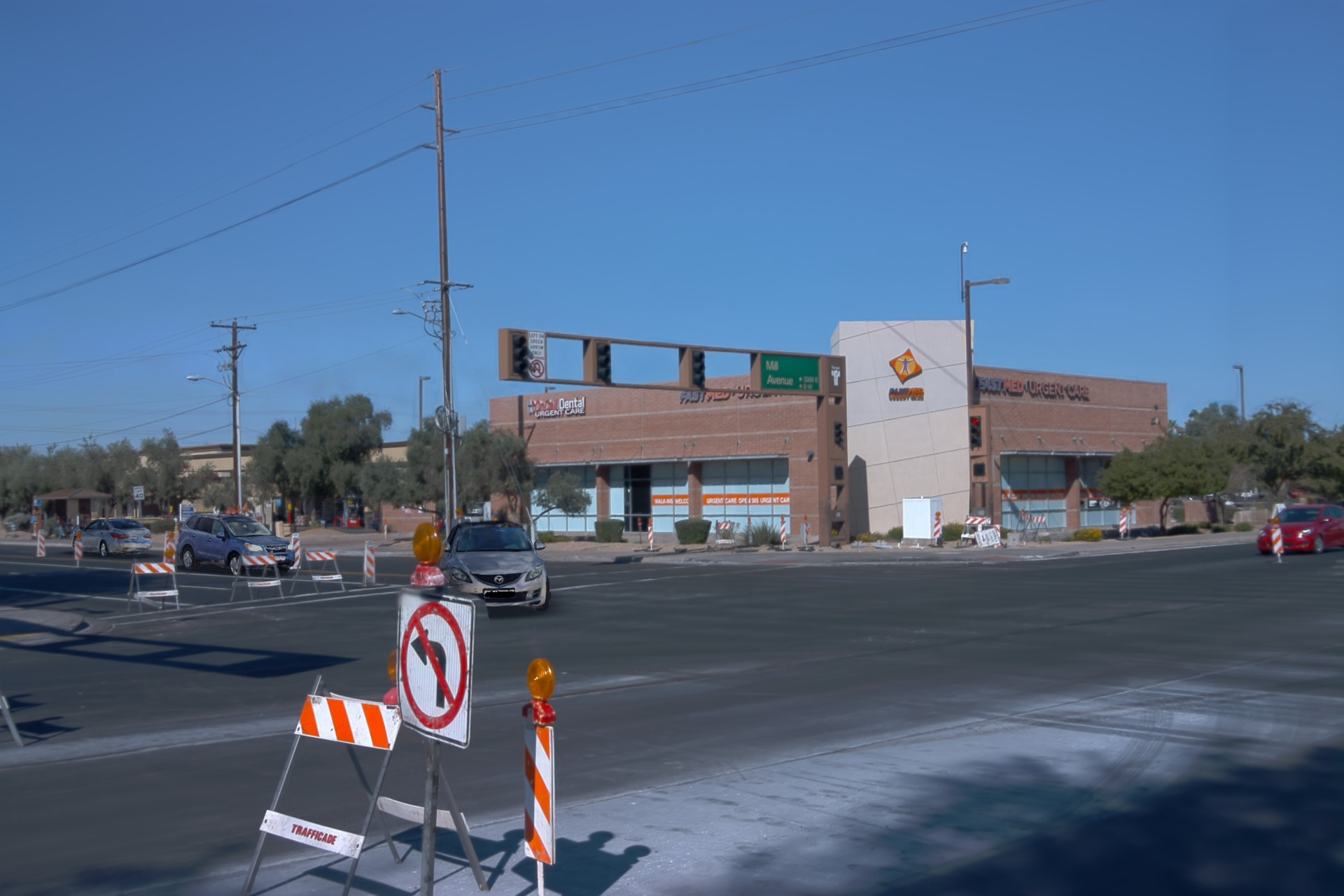} &
         \includegraphics[width=0.2\linewidth]{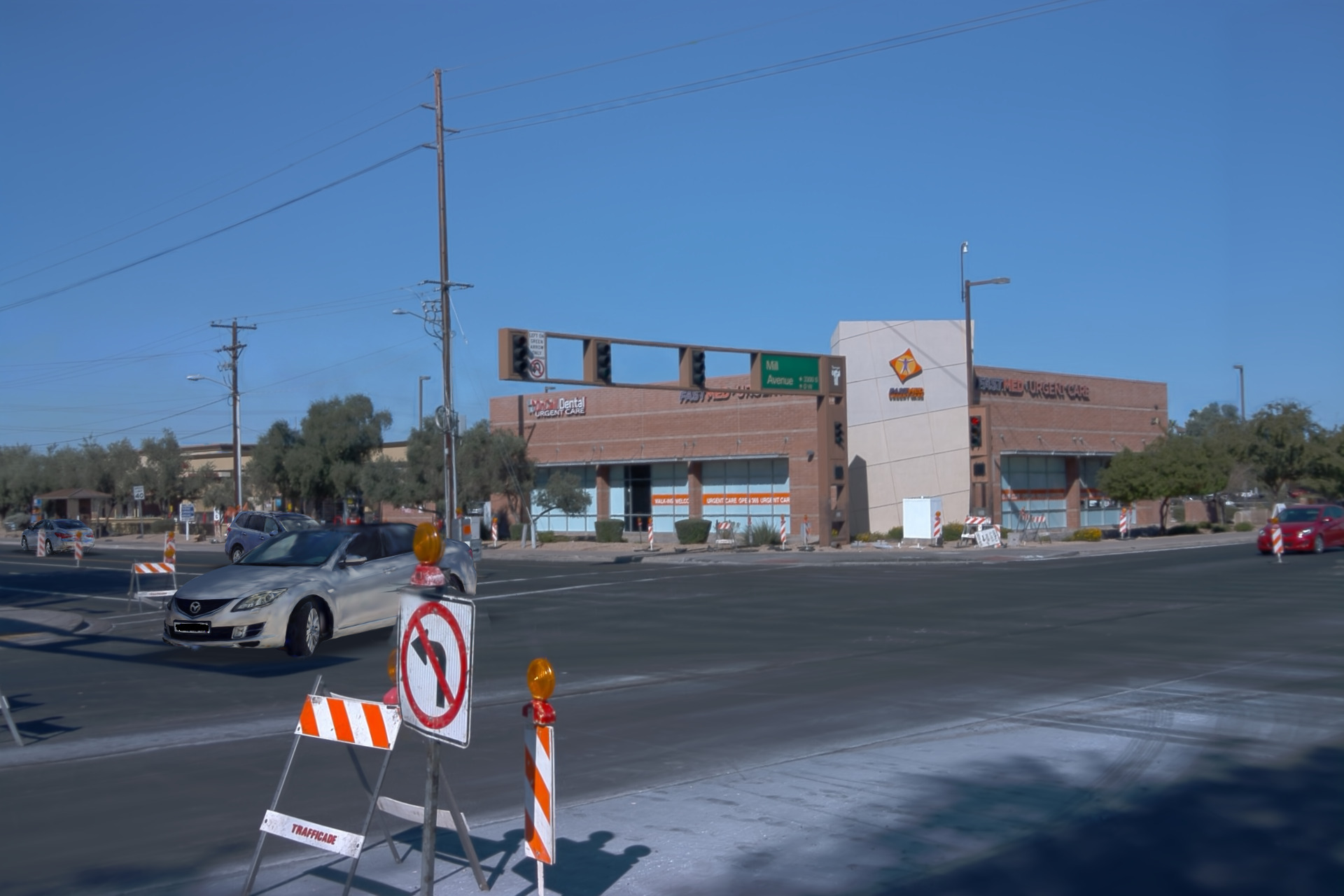} &
         \includegraphics[width=0.2\linewidth]{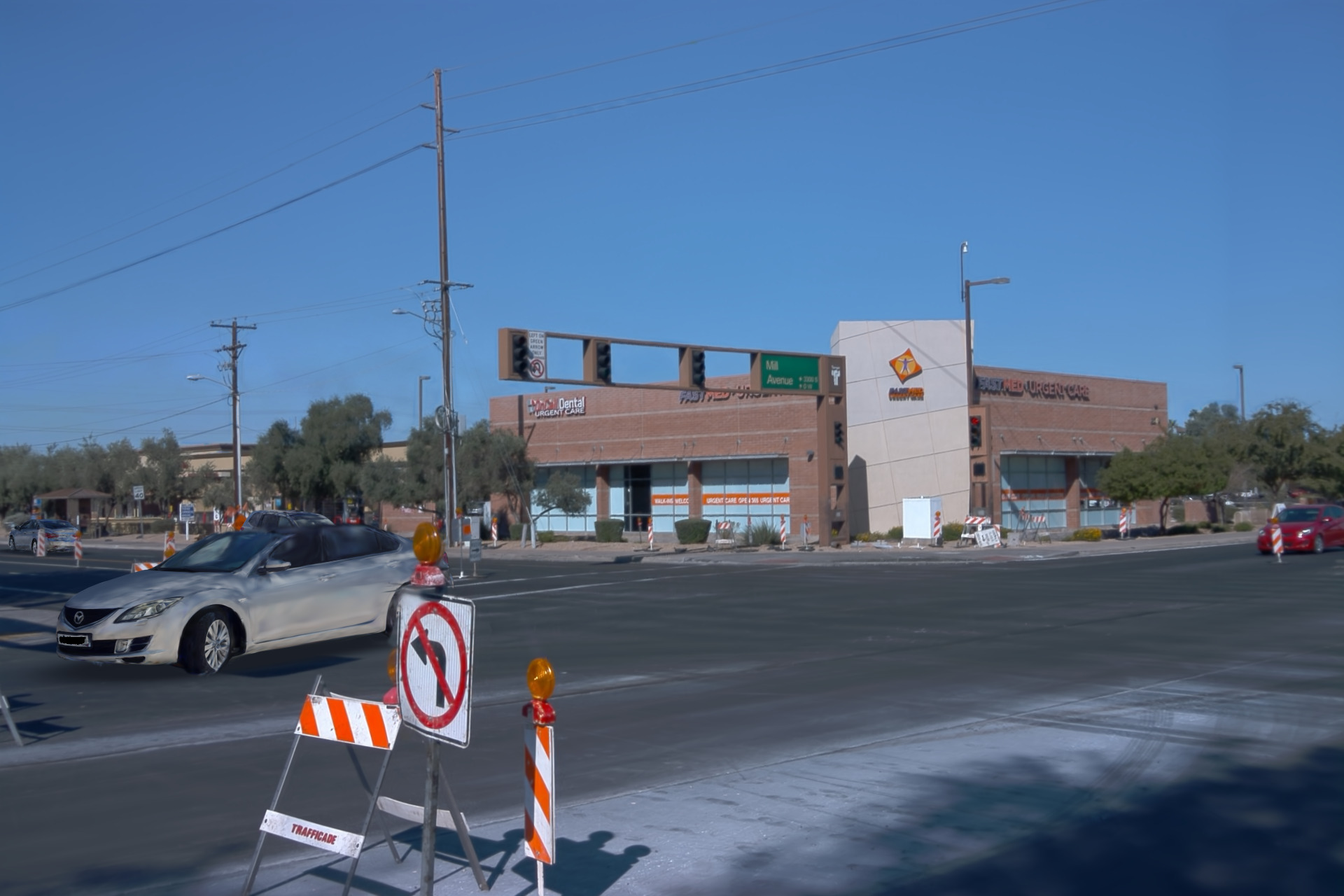} \\

     \end{tabular}
     \caption{Visualization of original and modified trajectories with \ourmethod. The cars retain high-fidelity appearance even at close distances to the ego camera.}
     \label{fig:results_grid}
\end{figure*}

We provide additional visual results in Figures \ref{fig:qualitative_frame_comparison_v2} and \ref{fig:qualitative_frame_comparison_v3}. We also demonstrate our method's capability to render novel views with substantial scene variations. \fig{results_grid}~showcases results across four test scenes, where all modifications preserve high image quality.

\section{Limitations}\label{appendix:limitation}




\paragraph{Reconstruction limitations.}
To run reconstruction, we estimate camera parameters from the input images.
In particular, we run bundle adjustment starting from the initialization obtained with VGGT.
At present, errors in camera estimation remain a primary source of reconstruction failures.
We expect that continued advances in foundational vision models will substantially reduce this limitation.

State-of-the-art multiview reconstruction methods continue to struggle with reflective and glossy surfaces like cars even up to this day.
Accurate modeling of reflections on metallic surfaces on real datasets demands more precise representations of illumination - beyond what conventional environment maps can provide.

\paragraph{Dataset limitations.}
We rely on an external dataset sourced from online car sale advertisements, which primarily features passenger vehicles.
As a result, other vehicle categories (e.g. buses, trucks, and service vehicles) are underrepresented and cannot yet be reliably replaced by our method.

Nonetheless, our pipeline is fully modular: adding support for additional vehicle types only requires capturing a 360° video of the target vehicle to add it to the retrieval database.

\section{Statement on LLM usage}
The authors used the large language model (LLM) only to improve the writing and grammar of the text. All the results from the LLM were checked by the authors.


\end{document}